\documentclass[sigconf, nonatbib]{acmart}
\renewcommand\footnotetextcopyrightpermission[1]{} % removes footnote with conference information in first column
\usepackage{fontawesome}
\usepackage{multirow}
\usepackage{arydshln}
\usepackage{comment}
\usepackage{subcaption}
\usepackage{booktabs}
\usepackage{cancel}
\usepackage{relsize}

\usepackage{pifont}
\usepackage{subcaption} % 子图支持
\definecolor{red}{RGB}{255, 60, 60}  
\definecolor{blue}{RGB}{120, 150, 250} 
\definecolor{green}{RGB}{80, 220, 80}

\begin{document}

%%
%% The 'title' command has an optional parameter,
%% allowing the author to define a 'short title' to be used in page headers.
\title{Towards Explainable Partial-AIGC Image Quality Assessment}
\author{Jiaying Qian\textsuperscript{1*}, Ziheng Jia\textsuperscript{1*}, \\ Zicheng Zhang$^1$, Zeyu Zhang$^1$,Guangtao Zhai$^1$, Xiongkuo Min$^{1\diamondsuit}$ \\
$^1$Shanghai Jiaotong University
}
%%
%% The 'author' command and its associated commands are used to define
%% 'authornote' and 'authornotemark' commands

\begin{abstract}
The rapid advancement of AI-driven visual generation technologies has catalyzed significant breakthroughs in image manipulation, particularly in achieving photorealistic localized editing effects on natural scene images (NSIs). Despite extensive research on image quality assessment (IQA) for AI-generated images (AGIs), most studies focus on fully AI-generated outputs (e.g., text-to-image generation), leaving the quality assessment of \textbf{partial-AIGC images} (PAIs)—images with localized AI-driven edits—an almost unprecedented field. Motivated by this gap, we construct the first large-scale PAI dataset towards \underline{e}xplainable \underline{p}artial-\underline{A}IGC \underline{i}mage \underline{q}uality \underline{a}ssessment (EPAIQA), the \textbf{EPAIQA-15K}, which includes $15$K images with localized AI manipulation in different regions and over $300$K multi-dimensional human ratings. Based on this, we leverage large multi-modal models (LMMs) and propose a three-stage model training paradigm. This paradigm progressively trains the LMM for editing region grounding, quantitative quality scoring, and quality explanation. Finally, we develop the \textbf{EPAIQA series models}, which possess  \textbf{explainable} quality feedback capabilities. Our work represents a pioneering effort in the perceptual IQA field for comprehensive PAI quality assessment.

\end{abstract}

\vspace{-20pt}



%% A 'teaser' image appears between the author and affiliation
\begin{teaserfigure}
    \centering
  \includegraphics[width=0.98\textwidth]{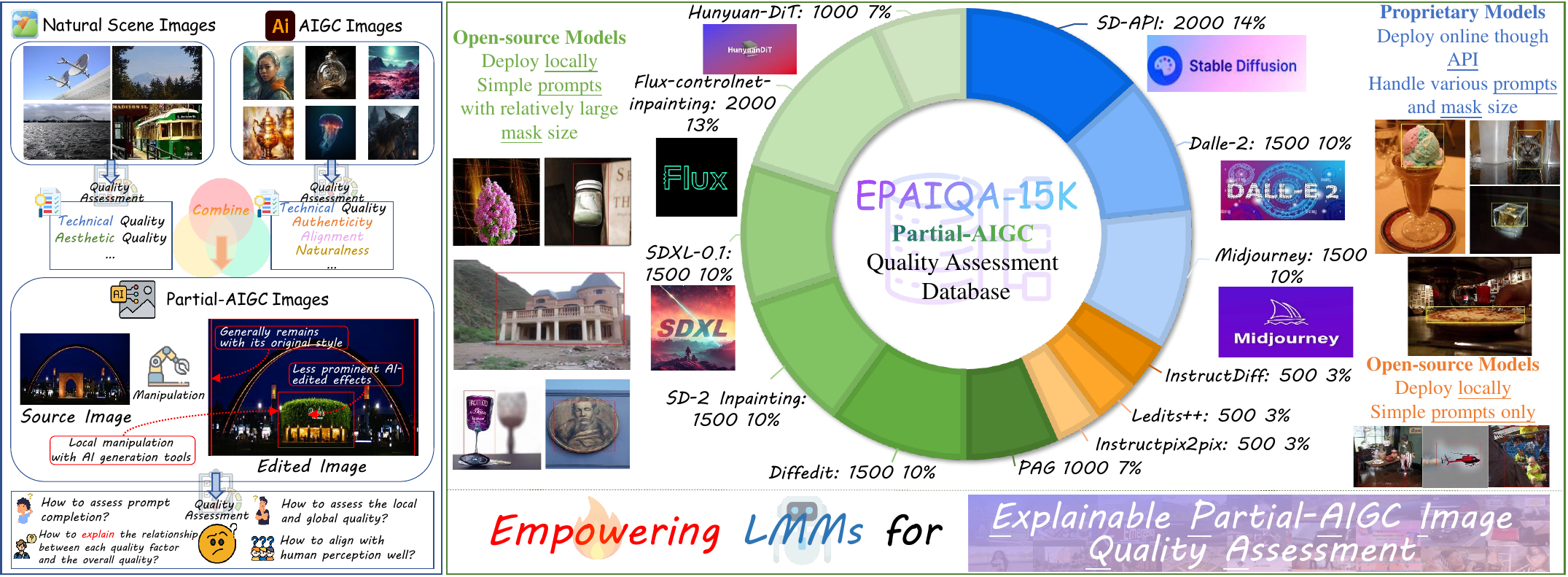}
  \vspace{-6pt}
  \caption{On the left is the motivation of our work: While substantial advancements have been achieved in quality assessment for NSIs and AGIs, research on the PAIQA remains almost unprecedented and faces significant challenges. On the right is an overview of our EPAIQA-15K dataset, which involves 12 different image editing tools, including open-source and proprietary models with different input requirements.}
  \vspace{-3pt}
  % \Description{Enjoying the baseball game from the third-base
  % seats. Ichiro Suzuki is preparing to bat.}
  \label{fig:spotlight}
\end{teaserfigure}

\maketitle

\let\thefootnote\relax\footnotetext{\noindent\textsuperscript{*}Equal contribution.  $^\diamondsuit$Corresponding author.}

\begin{figure*}[htbp]
  \centering
  \includegraphics[width=0.98\linewidth]{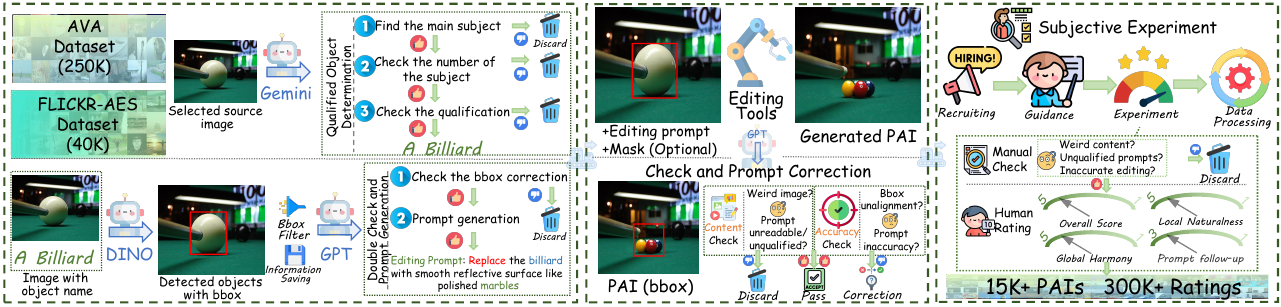}
  \vspace{-10pt}
  \caption{Data construction pipeline of the EPAIQA-15K dataset.}
  \vspace{-3pt}
   \label{pipeline1}

\end{figure*}

\section{Introduction}
Perceptual image quality assessment (IQA) constitutes a foundational area within the field of digital image processing, with extensive research dedicated to evaluating the quality of natural scene images (NSIs) spanning several decades \cite{wang2004image,zhai2020perceptual}. With the maturation and widespread adoption of AI-based visual generation technologies, research on AI-generated image quality assessment (AGIQA) has emerged and rapidly advanced over the past two years \cite{ImageReward, AGIQA-3K}. Nevertheless, the majority of these studies primarily concentrate on text-to-image (T2I) image quality assessment, as the images exhibit distinct characteristics of AI-generated content (AIGC) that significantly differ from typical NSIs in both statistical and rendering properties. In addition to T2I image generation, generative AI-based image manipulation (editing) has also experienced a boost in recent years. These techniques focus on localized editing, targeting specific objects in images (such as object addition, erase, replacement, enhancement, and style transfer). Their editing effects pursue authenticity and always remain confined to specified regions, thereby ensuring that the manipulated images \textbf{retain a considerable portion of the original natural scene while simultaneously incorporating less prominent localized AIGC characteristics}. We designate these images as \textbf{partial-AIGC images (PAIs)}. However, the perceptual PAI quality assessment (\textbf{PAIQA}) represents an unprecedented field, thereby presenting a potential space for further exploration. Here, we would like to throw out several questions:

%as the images exhibit distinct characteristics of AI-generated content (AIGC) that significantly differ from typical NSIs in both statistical and rendering properties.with the images displaying distinct characteristics of AI-generated content (AIGC) that differ markedly in both statistical and rendering properties from typical NSIs

\textit{What is the significance of PAI and PAIQA?}

PAIs are closer to NSIs compared to typical complete AGIs, thus making these images more capable of replacing NSIs that would otherwise require expensive shooting costs, offering broad \textbf{practical value}. PAIQA integrates assessments of the quality of localized AI editing effects and the image global harmony, thus can provide considerable \textbf{feedback} for image manipulation refinement.

\textit{What are the challenges for PAIQA models?}

% Firstly, compared to traditional no-reference (NR) IQA models, which only require single-image input, and most AIGC-IQA methods, which typically require only one
%  single image and prompt, 
Firstly, compared to  most traditional IQA and AGIQA methods, 
 PAIQA models involve more complex inputs. A complete evaluation of a manipulation sample necessitates at least the triplet of the \textbf{source image}, the \textbf{edited image}, and the \textbf{editing prompt}. This demands that the model be equipped with the fundamental capability to accept multi-stimulus, multi-modal inputs, while also possessing high-level semantic understanding and a local grounding ability. Furthermore, the quality factors of PAIQA span multiple perspectives from high-level to low-level. The former primarily focuses on the editing prompt completion. At the same time, its counterpart delves into the assessment of the edited region's local naturalness as well as the overall harmony of the entire image. Moreover, to comprehensively weigh these factors and derive an \underline{e}xplainable PAIQA (EPAIQA) feedback that possesses high alignment with human perception also presents a significant challenge.

In response to the aforementioned challenges, we introduce \textbf{EPAIQA-15K}, the \textbf{first} large-scale dataset specifically curated for comprehensive PAIQA, setting a new benchmark in the field. We adopt an elaborate data selection pipeline involving the collaboration of object detection models and large multi-modal models (LMMs). From over $250$K NSIs, we pick over $15$K diverse source images that include primary objects and generate their related editing prompts. Subsequently, we use $12$ image editing tools to perform local manipulation. We also conduct rigorous data filtering processes to maintain the quality of the dataset used for subjective experiments. The human evaluation subjective experiment encompasses four distinct quality dimensions, thereby ensuring a robust and sufficiently comprehensive dataset construction.

Considering the requirements for the model's foundational capabilities, we use LMMs as the base model for our EPAIQA model. The model training is divided into three stages. First, we perform pre-training centering on the editing region grounding task. This step enhances the LMM's sensitivity to the localized AI-edited effects. Next, we train the LMM to enhance the quantitative scoring capabilities of the edited image's local naturalness and overall harmony. Finally, we inject the chain of thought (\textbf{CoT}) reasoning knowledge to construct a fine-grained instruction tuning dataset. Through systematic model training, we obtain our final EPAIQA model, which is capable of providing \textbf{explainable} and nuanced quality feedback.

Our core contributions are three-fold:

\begin{enumerate}
    \item We build the first large-scale PAIQA dataset, the \textbf{EPAIQA-15K}, which incorporates a variety of \textbf{image contents}, \textbf{editing tools}, \textbf{manipulation tasks}, and \textbf{editing region sizes}. We also conduct strict data scrutiny and subjective experiments to ensure the quality and abundance of the dataset. 
    \item We design a three-stage LMM training process. For each stage, we construct rich instruction tuning data using various language-based input formats. 
    \item Finally, we develop the \textbf{EPAIQA series models}, which exhibit outstanding quality scoring and explainable quality feedback capabilities.
\end{enumerate}

\section{Related Works}
\subsection{Image Editing}
Image editing aims to perform manipulation on a specific region or object of a source image. In terms of manipulation types \cite{huang2024diffusion}, image editing includes \textbf{semantic editing}, including objects adding or removing, object replacement, and objects semantic changes; \textbf{stylistic editing}, like style transfer, lighting adjustments, color changes, and highlighting or enhancing specific objects; and \textbf{structural editing}, such as object movement and perspective transformation. Concerning input requirements (except for the source image), the models can be principally categorized into three types: those that necessitate a reference image, editing prompt, and editing region mask \cite{xie2023dreaminpainter,yang2023uni,lu2023tf}; those that rely on a prompt and mask \cite{li2024zone,huang2025pfb,yu2023inpaint,avrahami2023blended}; and those that operate solely on prompt inputs \cite{huberman2024edit,tumanyan2023plug,mirzaei2024watch}. 

% The mask types include binary masks and those that only erase the edited region while keeping the remaining areas original.

\begin{figure*}[htbp]
  \label{11}
  \centering
  \includegraphics[width=0.995\linewidth]{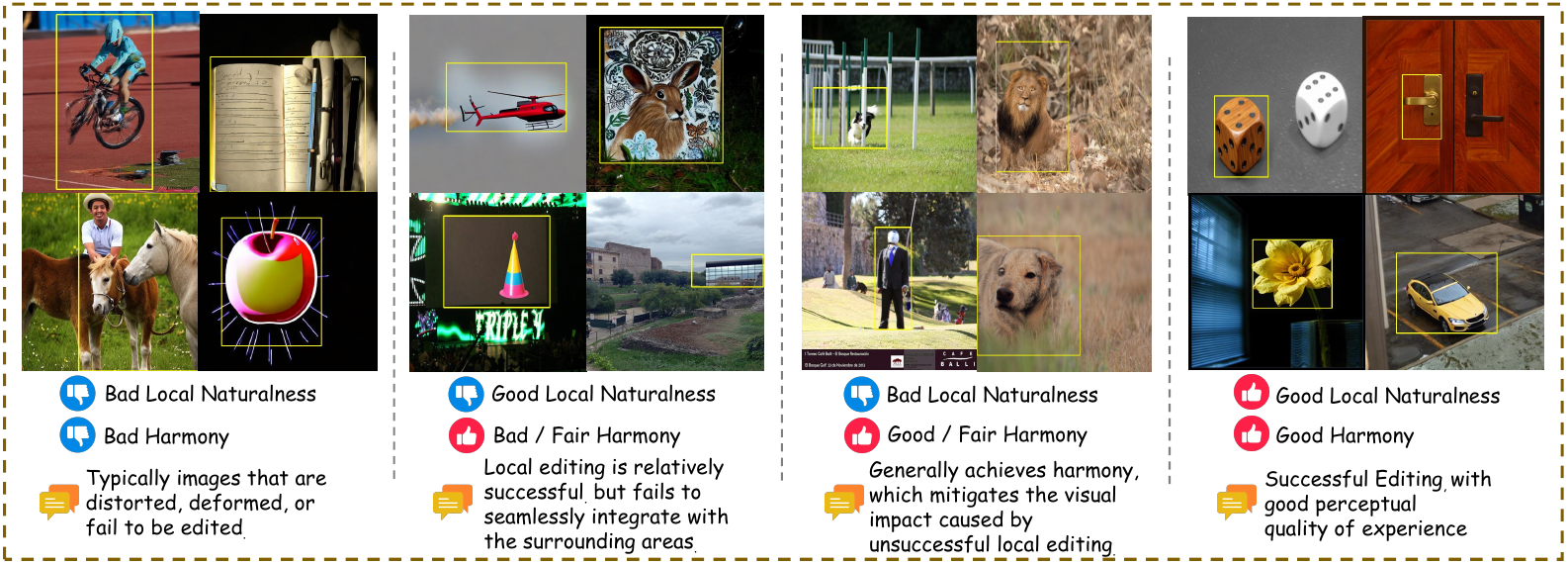}
  \vspace{-10pt}
  \caption{Examples of edited images with divergent harmony and local naturalness quality level.}
  \label{exp1}
\end{figure*}

\vspace{-8pt}
\subsection{Image Quality Assessment}

Traditional IQA methods can be mainly divided into \textbf{full-reference (FR)} IQA \cite{VIF,MAD,FSIM,GSM,VSI,GMSD} and \textbf{no-reference (NR)} IQA \cite{twostep,NSS,DCT,BRISQUE,NIQE}. Determined by model complexity, they can also be divided into statistics methods and deep neural network (DNN)-based methods. The former, such as \cite{SSIM,MS-SSIM,IW-SSIM}, directly assess the images' inherent statistical features or structural information.  For the latter, \cite{IQA-CNN,RankIQA, DBCNN,HyperIQA} apply convolutional neural networks (CNN) to assess image quality. With the emergence of transformers \cite{vaswani2017attention}, some researches \cite{MUSIQ,TRIQ,NTIRE} bring new architectures. In recent years, AGIQA \cite{IP-IQA,PickScore,IPCE,MoE-AGIQA,MA-AGIQA}  has become a new emerging trend. Researchers have also established a bunch of AGIQA datasets, such as DiffusionDB \cite{DiffusionDB}, AGIQA-$1$K \cite{AGIQA-1K}, ImageReward \cite{ImageReward}, Aigciqa2023 \cite{wang2023aigciqa2023}, AGIQA-$3$K \cite{AGIQA-3K}, and AGIQA-$20$K \cite{AGIQA-20K}. However, existing methods and datasets predominantly focus on complete AGIs, especially for the T2I images, which implies that no specialized research is currently available for PAIQA, thus leaving considerable exploration space. These gaps have become the motivations for our work.

\section{The EPAIQA-15K Dataset}
% \begin{figure*}[htbp]
%   \label{11}
%   \centering
%   \includegraphics[width=\linewidth]{samples/imgs/examples1.pdf}
%   \caption{Bounding Box Size Distribution}
% \end{figure*}
% \begin{figure}[htbp]
%   \label{11}
%   \centering
%   \includegraphics[width=0.8\linewidth]{samples/imgs/Cross.pdf}
%   \caption{Bounding Box Size Distribution}
% \end{figure}
The EPAIQA-$15$K comprises $15,026$ PAI data samples, with each sample including the quadruple of the source image, the edited image, the edit prompt, and the precise editing region coordinates. Editing examples  are shown in \textit{supplementary material}  Sec. \ref{APED:example}. In the dataset construction process, we implement a multi-stage filtering mechanism to ensure data quality and diversity. 

\subsection{Dataset Construction Pipeline}

The dataset construction pipeline is divided into three stages, as shown in Fig. \ref{pipeline1}.
% \begin{itemize}
%     \item \textbf{Image-Prompt Pair Generation} We select appropriate images from open-source aesthetic datasets and generate editing prompts covering four types of editing tasks: object operation, object enhancement, semantic change, and style change, to construct a diverse range of editing tasks.
% \end{itemize}
% \begin{itemize}
%     \item \textbf{Large-scale Automated Image Editing} We employ over ten image editing methods to perform local edits on images based on the editing prompts, and apply pre-established screening criteria to rigorously filter the generated data to ensure data quality.
% \end{itemize}
% \begin{itemize}
%     \item \textbf{Subjective Experiment} Participants were asked to evaluate the edited images based on the source images and editing prompts, providing scores for overall editing quality, harmony, naturalness of the edited regions, and prompt adherence.
% \end{itemize}
\begin{figure*}[htbp]
    \centering
    % \begin{subfigure}[t]{0.245\linewidth}
    %     \includegraphics[width=\linewidth]{samples/imgs/overall_score1.pdf}      
    %     \label{fig:sub1}
    % \end{subfigure}
    % \hfill
    % \begin{subfigure}[t]{0.245\linewidth}
    %     \includegraphics[width=\linewidth]{samples/imgs/harmony_score1.pdf}   
    %     \label{fig:sub2}
    % \end{subfigure}
    % \hfill
    % \begin{subfigure}[t]{0.245\linewidth}
    %     \includegraphics[width=\linewidth]{samples/imgs/local_naturalness1.pdf}
    %     \label{fig:sub3}
    % \end{subfigure} 
    % \hfill
    % \begin{subfigure}[t]{0.245\linewidth}
    %     \centering
    %     \includegraphics[width=0.8\linewidth]{samples/imgs/prompt_completion1.pdf}  
    %     \label{fig:sub4}
    % \end{subfigure}  
    % \par % 强制换行
    % % 添加垂直间距
    % \vspace*{-12pt} % 调整此处的值
     \begin{subfigure}[t]{0.245\linewidth}
        \includegraphics[width=\linewidth]{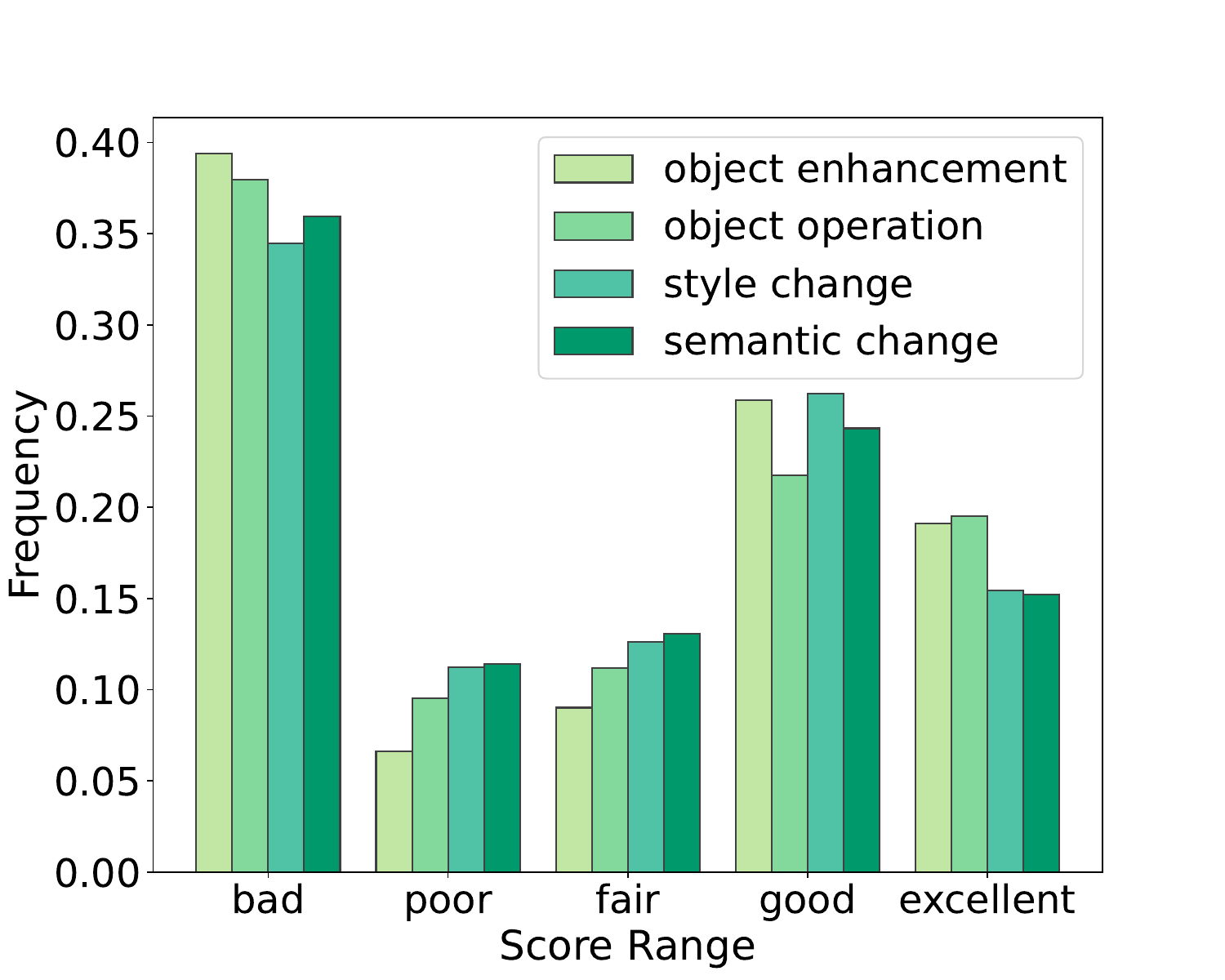}
        \caption{overall score}
        \label{fig:sub1}
    \end{subfigure}
    \hfill
    \begin{subfigure}[t]{0.245\linewidth}    
        \includegraphics[width=\linewidth]{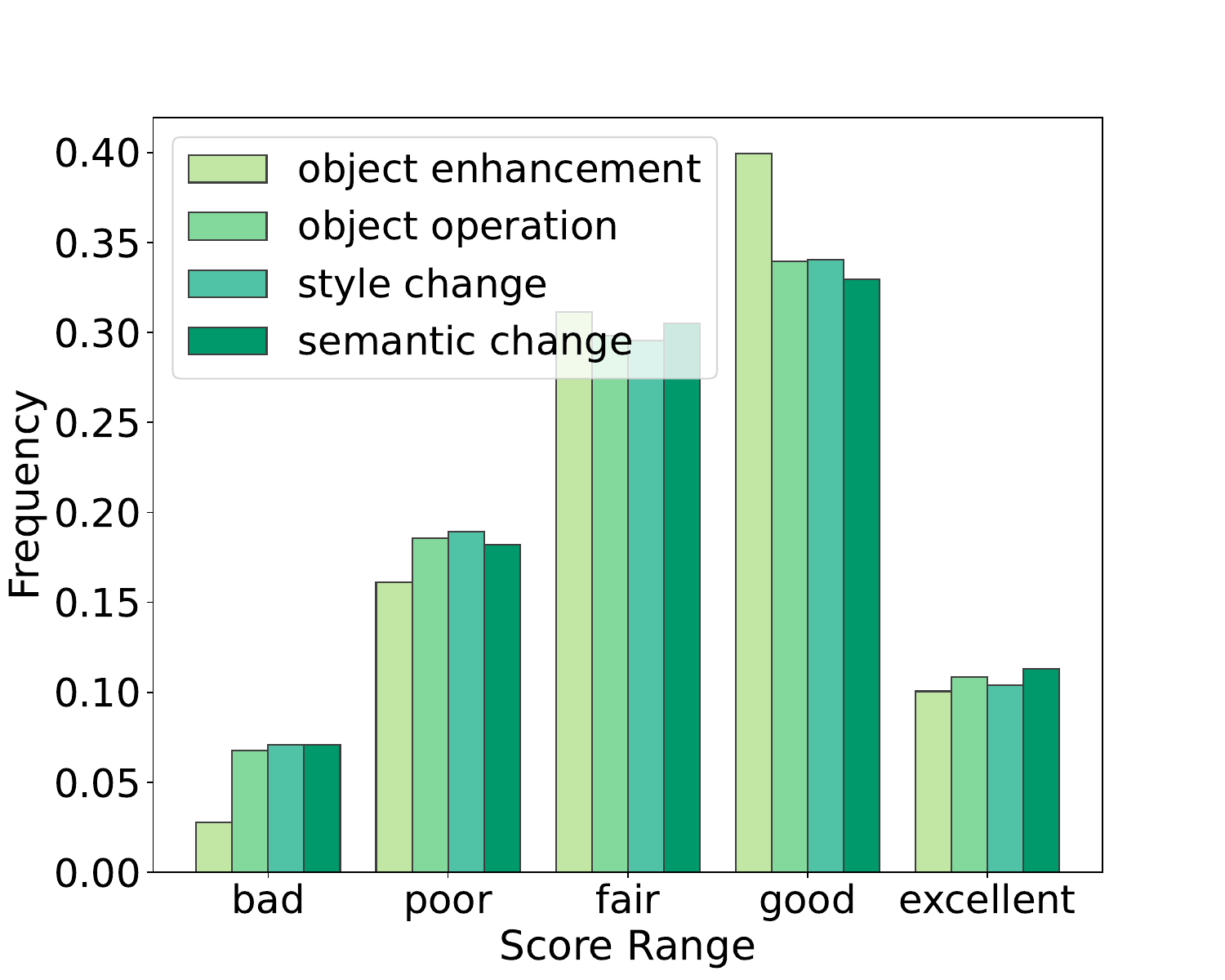}
        \caption{harmony score}
        \label{fig:sub2}
    \end{subfigure}
    \hfill
    \begin{subfigure}[t]{0.245\linewidth}   
        \includegraphics[width=\linewidth]{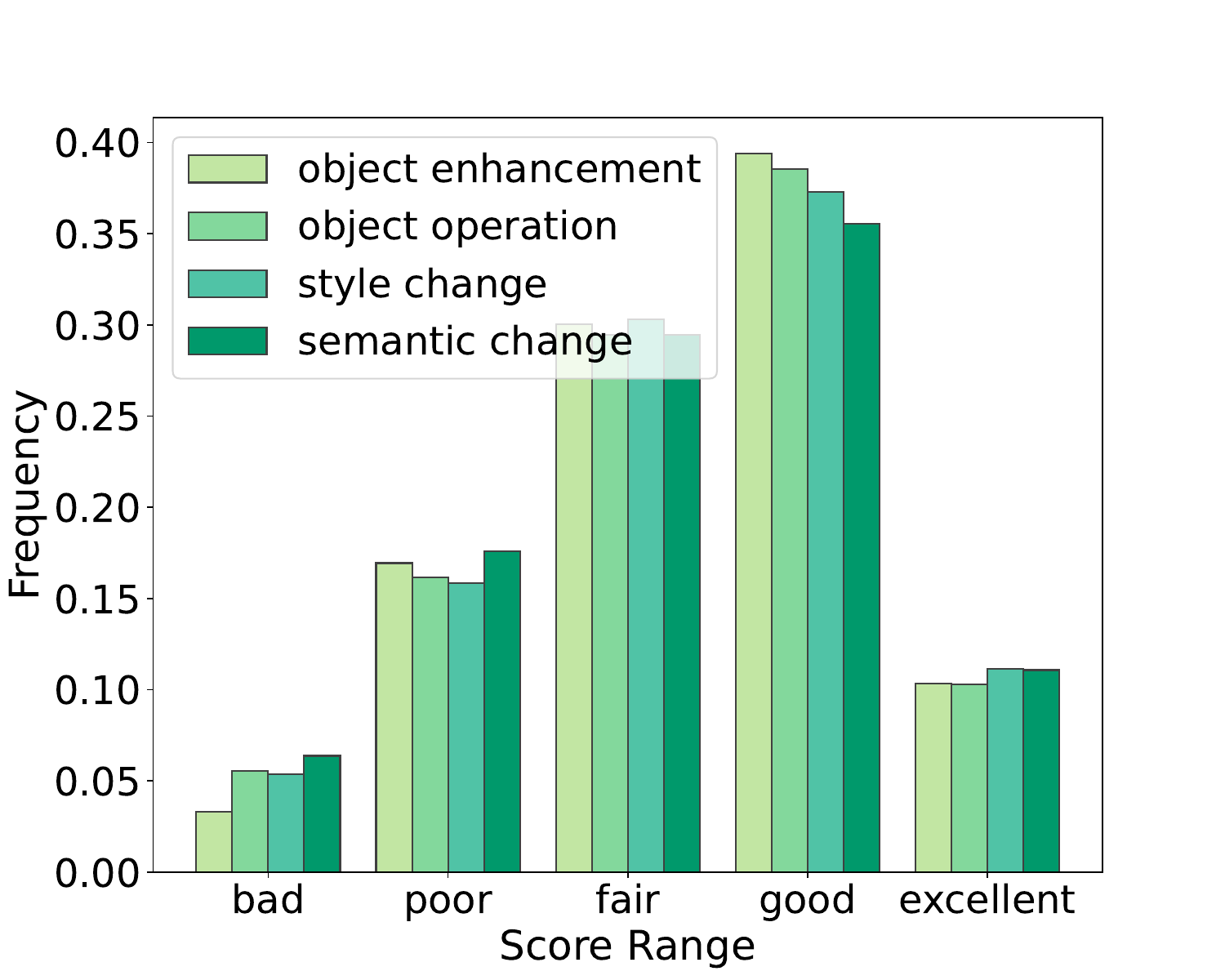}
        \caption{local naturalness}
        \label{fig:sub3}
    \end{subfigure}
    \hfill
    \begin{subfigure}[t]{0.245\linewidth}  
        \centering
        \includegraphics[width=0.8\linewidth]{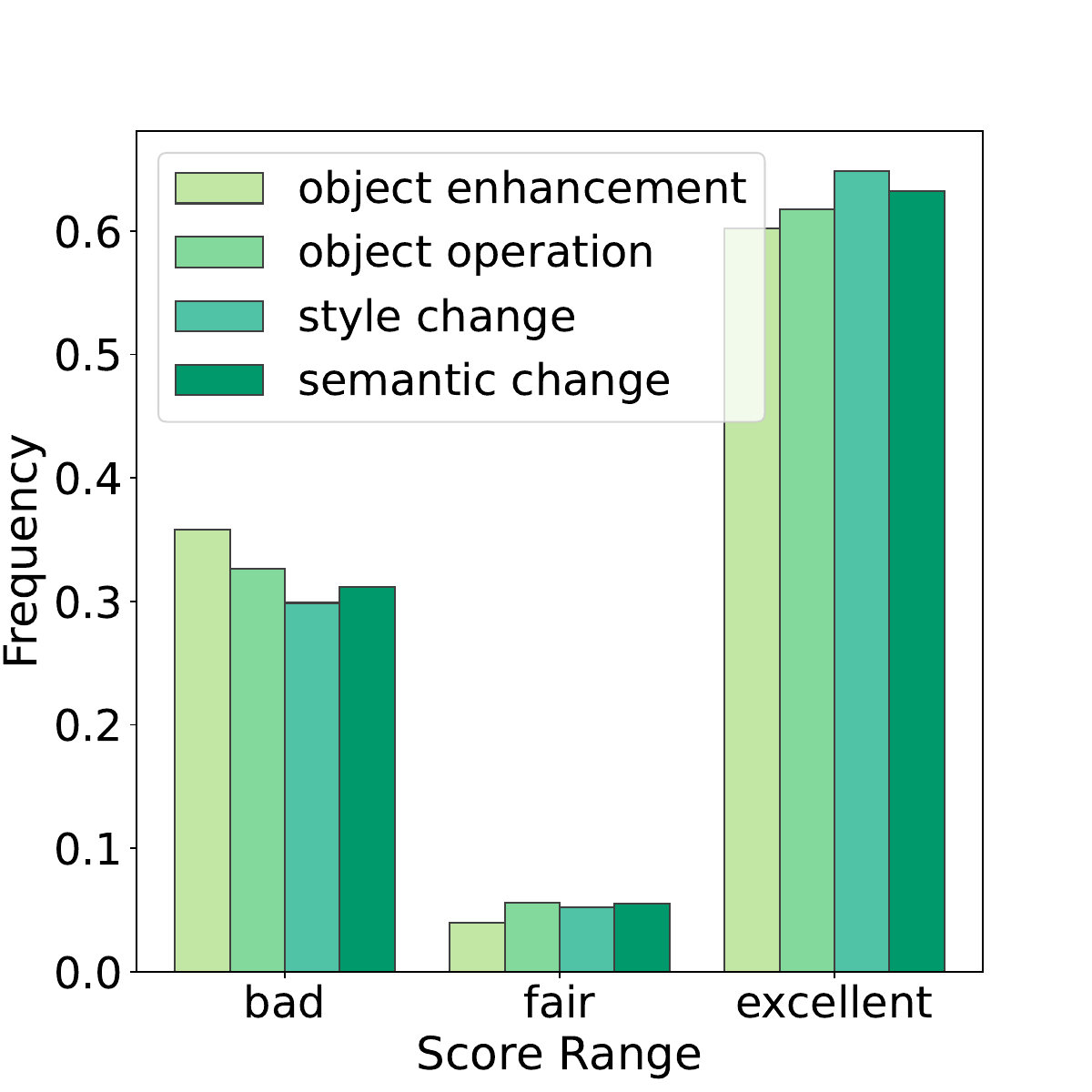}
        \caption{prompt completion}
        \label{fig:sub4}
    \end{subfigure}
    \vspace{-11pt}
    \caption{Human ratings distribution across four evaluation dimensions.}
    \label{fig:total1}
\end{figure*}

% \begin{figure*}[htbp]
%     \centering
%     \begin{subfigure}[b]{0.24\linewidth}
%         \includegraphics[width=\linewidth]{samples/imgs/overall_score2.pdf}
%         \caption{overall score}
%         \label{fig:sub1}
%     \end{subfigure}
%     \begin{subfigure}[b]{0.24\linewidth}
%         \includegraphics[width=\linewidth]{samples/imgs/harmony_score2.pdf}
%         \caption{harmony score}
%         \label{fig:sub2}
%     \end{subfigure}
%     \begin{subfigure}[b]{0.24\linewidth}
%         \includegraphics[width=\linewidth]{samples/imgs/local_naturalness2.pdf}
%         \caption{local naturalness}
%         \label{fig:sub3}
%     \end{subfigure}
%     \begin{subfigure}[b]{0.24\linewidth}
%         \includegraphics[width=\linewidth]{samples/imgs/prompt_completion2.pdf}
%         \caption{prompt completion}
%         \label{fig:sub4}
%     \end{subfigure}
%     \caption{Rating Distribution under the Four Editing Types}
%     \label{fig:total2}
% \end{figure*}

\noindent \textbf{Image-prompt pairs generation} 
We utilize publicly available image aesthetic datasets: the AVA Dataset \cite{AVA} and the FLICKR-AES Dataset \cite{FLICKR} as the foundation for our source images. The images in these datasets typically exhibit diverse content and well-designed compositions with relatively distinct main subjects, while they are generally free of significant technical distortions. This ensures data diversity and the controlability for subsequent image editing. 

We then leverage \textit{Gemini-1.5-pro} \cite{Gemini} for subject recognition, followed by a filtering process. First, \textit{Gemini} identifies the main objects in the images and their quantities (the prompts are in \textit{supp.} Sec. \ref{Gemini}). We select images with \textbf{only one main subject} and further remove images with main subjects of ambiguous semantics (such as light, ripples, etc.). Subsequently, we input the qualified images (with prompts) into the \textit{Dino} \cite{Dino} to detect the main objects, generating images with bboxes. Then, a second round of filtering is conducted to discard images with unqualified sizes of bboxes, retaining only those that have bbox sizes between \textbf{$5\%$} and \textbf{$75\%$} of the image area and with a reasonable width-height ratio. We then input the candidate images with bboxes into \textit{GPT-4o} \cite{GPT} to generate local editing prompts (prompts given to \textit{GPT} are shown in \textit{supp.} Sec. \ref{editing prompt}). Based on a systematic survey of existing editing models, we finally determine four editing tasks that are suitable for the subjective experiment: object operation, object enhancement, semantic change, and style change. 

% Among them, object operation includes erase and addition, while object enhancement aims to improve the prominence of the subject in the image. 

\noindent \textbf{Automated image editing}
We implement $12$ methods suitable for the editing process, which are 
\textit{Stable Diffusion API} \cite{SD}, \textit{DALLE-2} \cite{DALLE-2}, \textit{MidJourney}, \textit{InstructDiffusion} \cite{InstructDiff}, \textit{Ledits++} \cite{Ledits++}, \textit{Instructpix2pix} \cite{Instructpix2pix}, \textit{PAG} \cite{PAG}, \textit{Diffedit} \cite{Diffedit}, \textit{SD-2-inpainting}, \textit{SDXL-0.1-inpainting}, \textit{Flux-controlnet-inpainting} \cite{lipmanflow}, and \textit{Hunyuan-DiT-inpainting} \cite{Hunyuan-dit}. The number and proportion of edited images for each model are presented in Fig. \ref{fig:spotlight}. To ensure a high editing success rate, we assign more challenging prompts and images with smaller bboxes ($5\%$- $30\%$) to the proprietary models, while easier editing tasks and images with larger bboxes ($30\%$ -- $75\%$) are handled by locally deployed models, particularly text-only models.
\begin{figure}[htbp]
    \centering
    \begin{subfigure}[b]{0.46\linewidth}     
        \includegraphics[width=\linewidth]{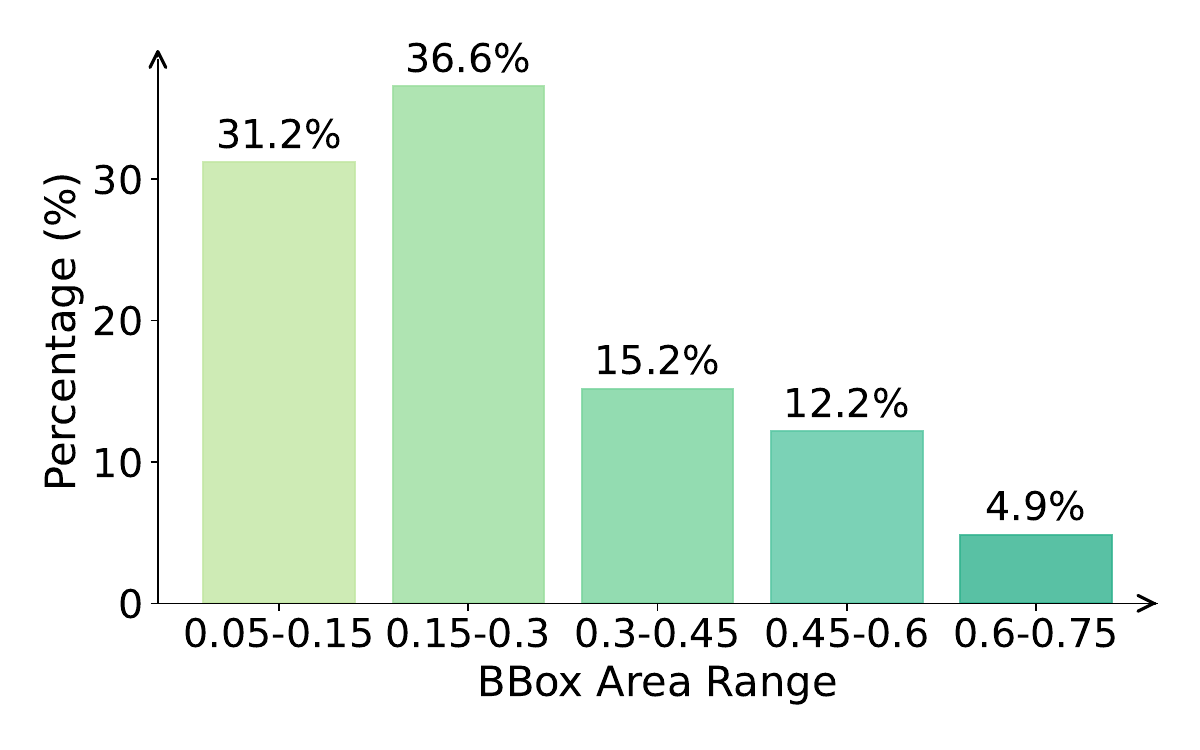}
        \caption{Bounding box size}
        \label{fig:total2sub1}
    \end{subfigure}
    \begin{subfigure}[b]{0.53\linewidth}
        \centering
        \includegraphics[width=0.8\linewidth]{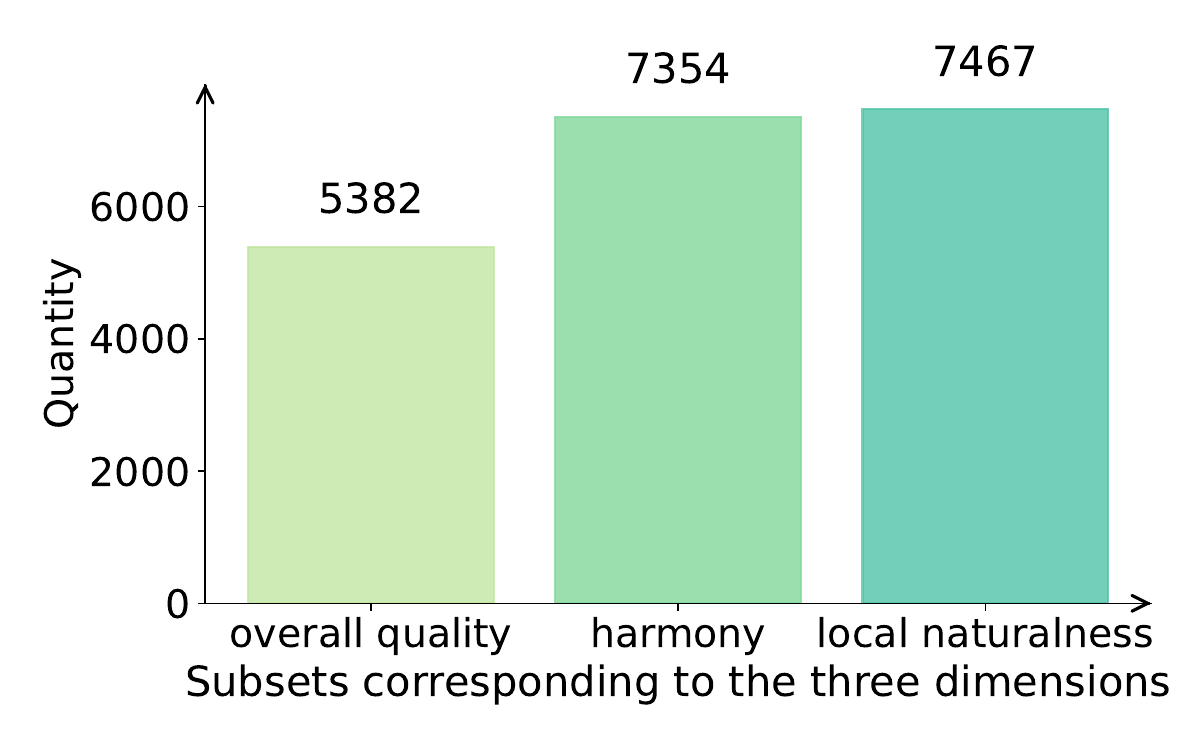}
        \caption{\# samples in three subsets}
        \label{fig:total2sub2}
    \end{subfigure}
    \vspace{-19pt}
    \caption{More information on data distribution}
    \label{fig:total2}
\end{figure}
% During the data generation process, we maintain a relatively stable number of edits from each model while slightly increasing the proportion of edited images generated by mainstream models. This strategy is designed to enhance the performance of mainstream models while preserving the overall robustness of the dataset, ensuring a balanced and high-quality data collection. 

% Before editing, the images are preprocessed to meet the respective input requirements of the models. 

Then, we integrate the source image, the binary mask (if needed), and the prompt for the editing model, which generates the edited image. To ensure the reliability of the manipulated images, we establish a strict data scrutiny mechanism using \textit{GPT} (shown in the middle-down part in Fig. \ref{pipeline1}). Firstly, the source images and edited images with visual anomalies are directly excluded. Then, editing prompts that are semantically ambiguous, unreasonable, or non-executable are eliminated. Finally, a prompt-subject alignment check is performed to remove edited samples in which the object required to be edited does not match the boxed one. The samples that meet all the scrutiny standards are used as the candidate images for the subjective experiment.

\noindent \textbf{Subjective experiment}
We have conducted a multi-dimensional subjective experiment and established rigorous settings and guidelines to enhance the reliability of the human-labeled results, which are elaborated in \textit{supp.} Sec. \ref{experiment}. The quality rating includes the following four key dimensions (editing examples with divergent subdimension quality levels are shown in Fig. \ref{exp1}) :
\begin{itemize}
\item \textbf{Overall editing quality} ($1$-$5$): This score is determined by both the \textbf{prompt completion} (with higher priority) and the \textbf{perception visual quality}. Scores of $1$-$2$ are assigned if (not only if) the editing fails to fully adhere to the prompt. 
\item \textbf{Harmony} ($1$-$5$): This dimension is independent of the prompt and focuses solely on the \textbf{style} and \textbf{semantic consistency} between the edited region and the rest of the image.
\item \textbf{Local naturalness} ($1$-$5$): It evaluates the naturalness of the edited region while also being decoupled from the prompt, focusing exclusively on the edited area.
\item \textbf{Prompt completion} ($1$-$3$): This dimension measures the prompt alignment ($1$: non-compliant, $2$: partly-compliant, $3$: fully compliant). 
\end{itemize}

We select $9,132$ images, with each image rated independently by at least $10$ participants. Participants simultaneously perceive the triplet of the source image, edited image, and prompt to determine their four-dimensional ratings. Any unreasonable samples, such as images still with visual abnormality passed during machine filtering, are manually excluded here.

After collecting the human-labeled data, we perform systematic cleaning by discarding ratings that don't meet the experimental criteria, such as the conflicting prompt completion scores and overall editing quality scores. We use the Interquartile Range (IQR) method to remove outliers in overall quality, local naturalness, and harmony scores. If an overall quality rating in one sample is excluded, its corresponding prompt completion score is also removed. The final score for the three dimensions for each editing sample is the mean of the qualified ratings, and the final prompt completion level is determined using a simple voting method.

% Through the aforementioned pipeline, we construct a comprehensive dataset comprising about $15,000$ PAIs and more than $300,000$ subjective ratings. 
% \vspace{-2pt}
\subsection{Dataset split}
\label{split}
After rigorous data screening, the final dataset contains a varying number of scored samples for each quality dimension. We construct three distinct subsets for further experiments. Firstly, for local naturalness and harmony, we curate the \textbf{naturalness and harmony subsets}, where each sample corresponds to an edited image (without bbox) along with its respective naturalness/harmony score. Subsequently, we develop the \textbf{overall quality subset}, and each sample includes the source and edited image pairs (without bboxes), the editing prompt, and the scores across all four dimensions. This implies that if a sample contains a quality dimension in which the score is filtered out before, then this sample is directly excluded from this subset. The number of samples in each subset is shown in Fig. \ref{fig:total2sub2}. Finally, we randomly partition the three subsets into non-overlapping training and testing sets using an $80\%/20\%$ split, while \textbf{ensuring that the training sets of the three subsets do not overlap with each other's test sets}. 

% \begin{figure}[t]
  
%   \centering
%   \includegraphics[width=0.98\linewidth]{samples/imgs/scatter.pdf}
%   \vspace{-5pt}
%   \caption{$3$D scatter plot of data across three dimensions}
%   \label{scatter}
% \end{figure}
\vspace{-12pt}
\subsection{Dataset Statistical Analysis}

\noindent \textbf{Data distributions} 
% \begin{figure}[htbp]
%   \label{11}
%   \centering
%   \includegraphics[width=\linewidth]{samples/imgs/bbox.pdf}
%   \caption{Bounding Box Size Distribution}
% \end{figure}
 The rating distribution for each dimension across the four editing types in EPAIQA-$15$K is shown in Fig. \ref{fig:total1}. The size (ratio) distribution of the bboxes is presented in Fig. \ref{fig:total2sub1}.  The ratings for each dimension and editing task  have no significant deviations or anomalous fluctuations, reflecting relatively high data quality and consistency. 

% As shown in Fig. \ref{11}, the size (ratio) distribution of the subject object bounding boxes covers the range from $0.15$ to $0.75$ of the total image area, with the data samples being most concentrated in the $0.3$ to $0.45$ range. This distribution aligns with the visual perception pattern, where objects of moderate size tend to be more common and visually perceivable.

\begin{figure}[t]
  
  \centering
  \includegraphics[width=0.95\linewidth]{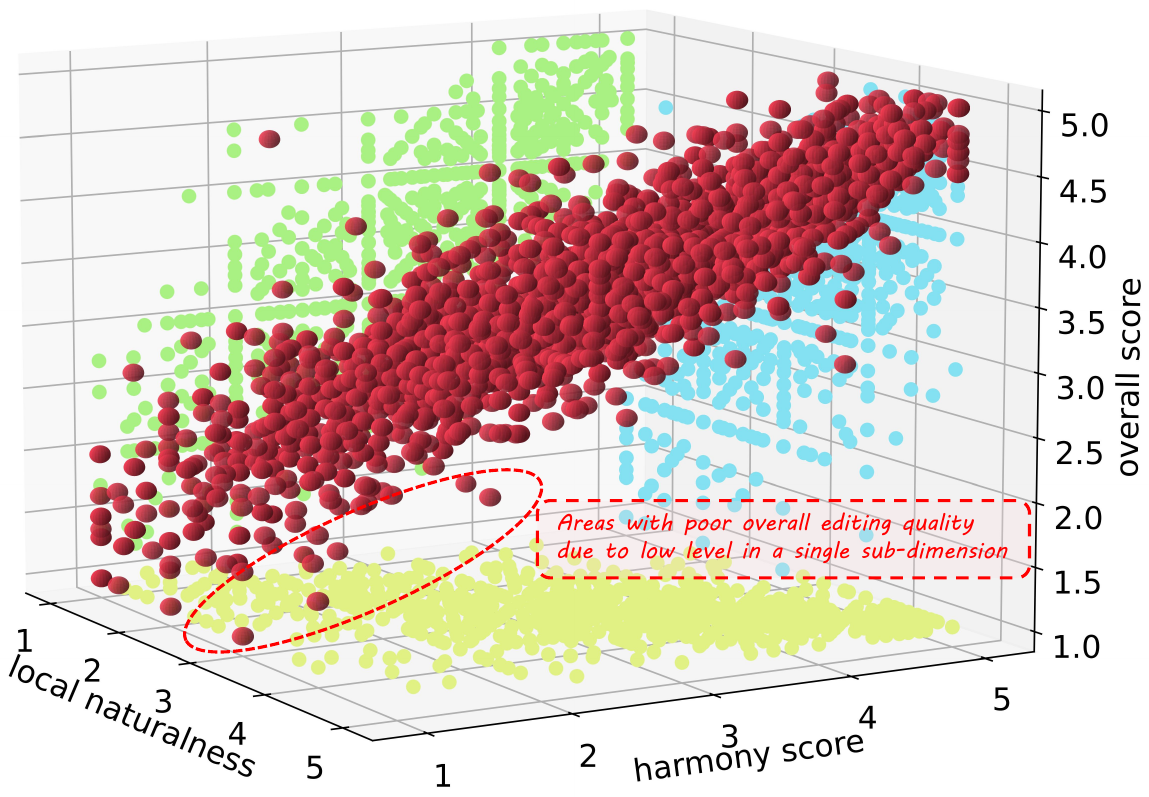}
  \vspace{-6pt}
  \caption{$3$D scatter plot of data across three dimensions}
  \label{scatter}
\end{figure}

\begin{figure}[t]
 
  \centering
  \includegraphics[width=0.95\linewidth]{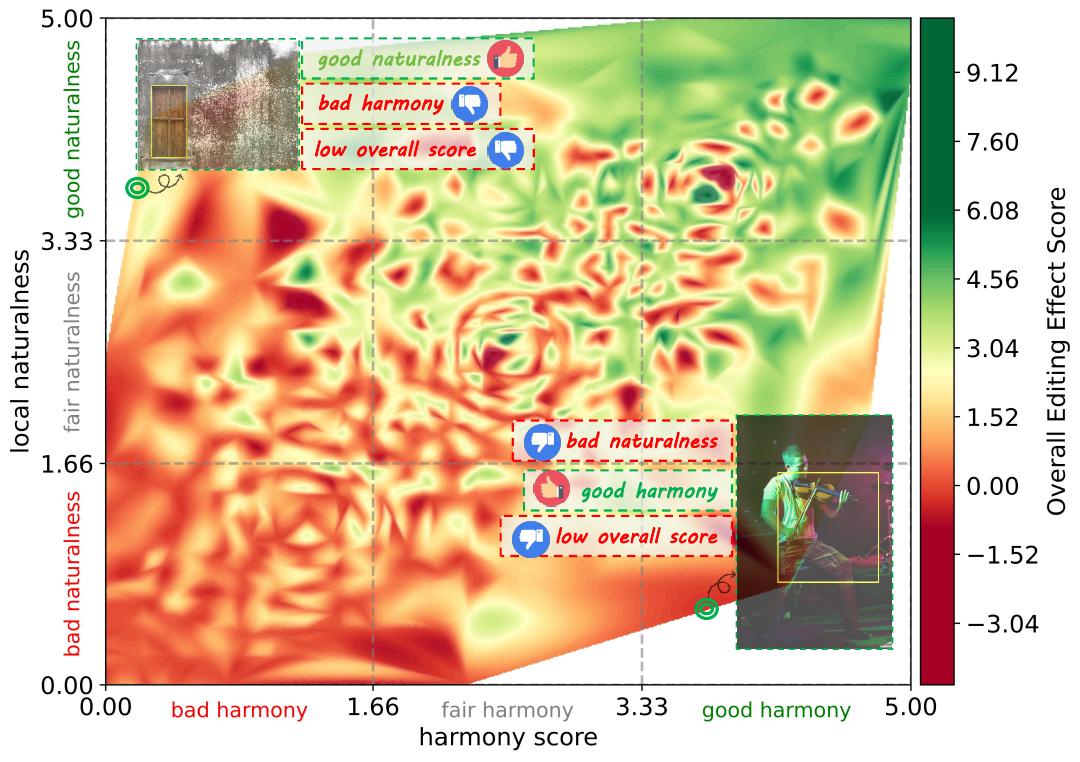}
  \vspace{-8pt}
  \caption{Divergence analysis heat map of harmony, local naturalness, and overall quality score.}
   \label{heat}
\end{figure}

\noindent \textbf{Data analysis} 
We also construct a 3D scatter plot to visualize the relationships among overall editing quality, harmony, and local naturalness (only using samples with full prompt completion), as illustrated in Fig. \ref{scatter}. To further investigate the relationship between quality dimensions, we visualize the divergence map of overall editing quality across the other two dimensions, as depicted in Fig. \ref{heat} (shown in a heat map style). 

We can observe from Figs. \ref{scatter} and \ref{heat} that under the condition of high-level prompt completion, there is, in general, a positive linear correlation across all dimension pairs.  Moreover, the linear relationship between the overall quality and harmony is more evident than that with local naturalness, suggesting that the overall editing effect may place greater emphasis on global consistency. It is particularly noteworthy that when either harmony or naturalness is at a low level, it more frequently precipitates a substantial decline in the overall score, even when the other sub-dimension does not demonstrate a low level. This is evidenced by the increased presence of outliers deviating from the linear relationship in the low-quality range of the overall score (highlighted by the \textcolor{red}{red box} in Fig. \ref{scatter}). This suggests that decoupling the analysis of editing effects related to local naturalness or global harmony facilitates the accuracy and reasonableness of the overall quality assessment.

\section{The EPAIQA Series Models}
We propose the EPAIQA series models based on supervised fine-tuning (SFT), with the three-stage training process shown in Fig. \ref{model}. The following is the elaboration of the three stages. All the prompts for machine annotation and formats of the instruction question-answer pairs are shown in \textit{supp.} Sec. \ref{modeltra}.
% \begin{itemize}
%     \item \textbf{Stage-1: Grounding Pre-training.} This stage focuses on equipping the model with the capability to accurately localize the target editing region based on the given prompt through large-scale data training, thereby establishing a foundation for precise region-specific quality assessment.

%     \item \textbf{Stage-2: Harmony and Local Naturalness Level Prediction Training.} In this phase, the model is trained to evaluate the visual harmony between the edited region and its surrounding context, as well as the perceptual naturalness of localized editing outcomes.

%     \item \textbf{Stage-3: Explainable Quality Assessment Training.} This final stage integrates the Chain of Thought (CoT) framework, enabling the model to not only score the overall editing quality but also generate detailed explanations and reasoning processes, thereby enhancing its interpretability and practical applicability in real-world scenarios.
% \end{itemize}

\subsection{Grounding Pre-training}
% \begin{figure}[t]
 
%   \centering
%   \includegraphics[width=0.95\linewidth]{samples/imgs/heat_map.pdf}
%   \vspace{-8pt}
%   \caption{Divergence analysis heat map of harmony, local naturalness, and overall quality score.}
%    \label{heat}
% \end{figure}

% \begin{figure*}[htbp]
%   \centering
%   \includegraphics[width=0.97\linewidth]{samples/imgs/model.pdf}
%   \vspace{-8pt}
%   \caption{The three-stage training pipeline for EPAIQA series models.}
%    \label{model}
% \end{figure*}

Stage-1 is the grounding pre-training process (shown in the left part of Fig. \ref{model}). We use the source image, the edited image, and the prompt as the training inputs. The LMM outputs four coefficients for the editing region, formatted as $\langle x, y, z_1, z_2 \rangle$, where $\langle x, y \rangle$ represents the center coordinates of the editing region, and $\langle z_1, z_2 \rangle$ denote the width and height of the region, scaled to $[0,1]$ to represent the ratio rather than absolute size. We conduct the pre-training using the typical language loss (such as GPT-loss \cite{radford2019language}) of the predicted parameters $\langle x, y, z_1, z_2 \rangle$ and the true parameters $\langle x_{\text{true}}, y_{\text{true}}, z_{1_{\text{true}}}, z_{2_{\text{true}}} \rangle$. This approach enables a more nuanced perception of localized editing effects. We utilize all the images from the EPAIQA-$15$K dataset, \textbf{excluding} the test sets of the three subsets designed in Sec. \ref{split}, for the instruction set in this stage. We then conduct the pre-training to obtain the stage-1 model.

% \begin{figure}[t]
 
%   \centering
%   \includegraphics[width=0.95\linewidth]{samples/imgs/heat_map.pdf}
%   \vspace{-8pt}
%   \caption{Divergence analysis heat map of harmony, local naturalness, and overall quality score.}
%    \label{heat}
% \end{figure}

\subsection{Quantitative Prediction Training}
\begin{figure*}[htbp]
  \centering
  \includegraphics[width=0.99\linewidth]{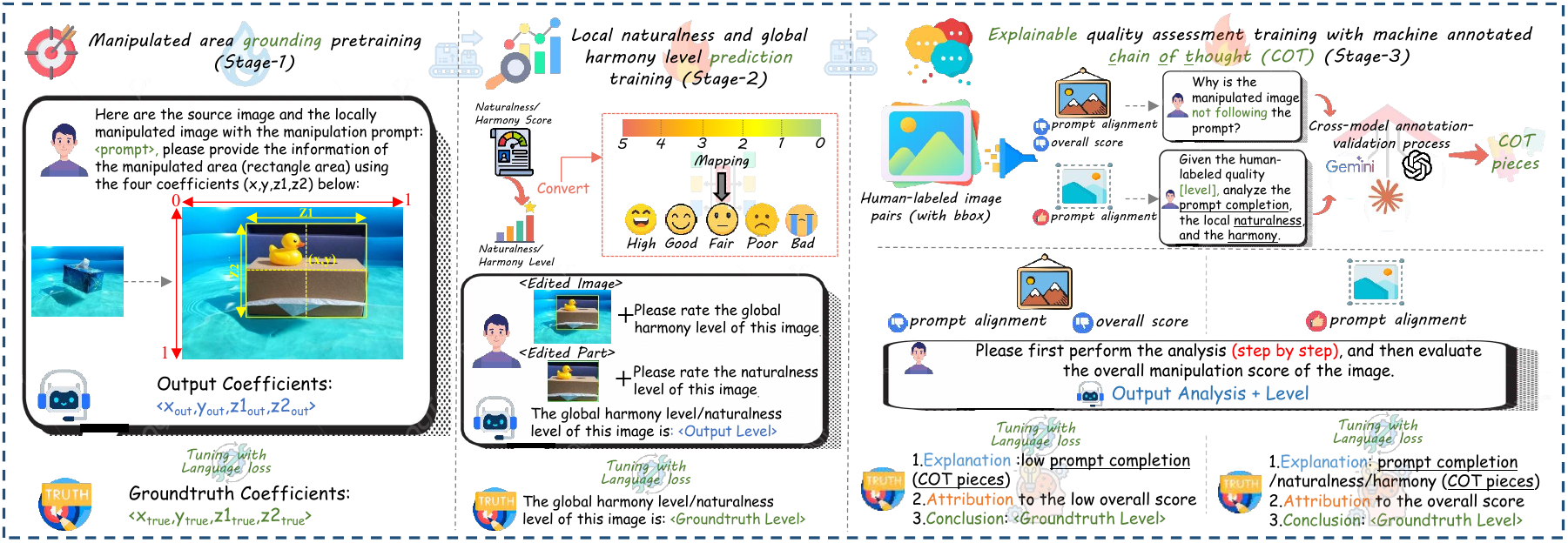}
  \vspace{-8pt}
  \caption{The three-stage training pipeline for EPAIQA series models. For clarity, ``Excellent'' is represented as ``High''.}
   \label{model}
\end{figure*}

Stage-2 concentrates on quantitative sub-dimension score prediction (shown in the middle part of Fig. \ref{model}). Following the strategy in \cite{qalign}, we first scale the data in the harmony and local naturalness subsets to $[0,5)$. Then we map these values to five quality levels: $[0,1)$: `bad', $[1,2)$:  `poor', $[2,3)$: `fair', $[3,4)$; `good', and $[4,5)$: `excellent'. 
In harmony prediction training, the entire edited image is fed into the model. In contrast, for local naturalness prediction training, only the cropped edited region (the boxed area) is provided as input to the model.
The data from both tasks are \textbf{mixed} to obtain a \textbf{unified} stage-2 model. We conduct the SFT process using the language loss between the model's output and the ground truth quality level based on the stage-1 model. Through these processes, the model progressively improves its ability to make quantitative quality predictions, building the foundation for the explainable quality understanding of PAIs. 

\begin{figure}[htbp]
  \label{11}
  \centering
  \includegraphics[width=0.99\linewidth]{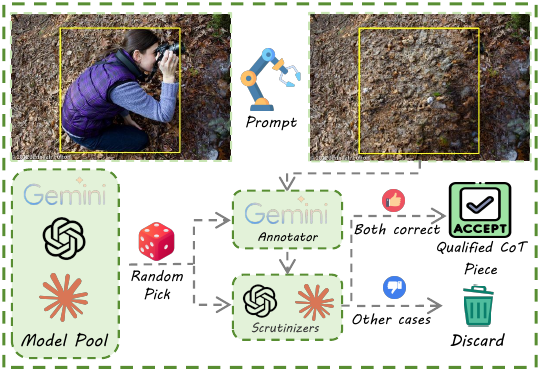}
  \vspace{-8pt}
  \caption{The cross-model annotation-validation process.}
  \label{cross}
\end{figure}

% \begin{figure*}[htbp]
%   \centering
%   \includegraphics[width=0.97\linewidth]{samples/imgs/model.pdf}
%   \vspace{-8pt}
%   \caption{The three-stage training pipeline for EPAIQA series models.}
%    \label{model}
% \end{figure*}

% \begin{figure}[htbp]
%   \label{11}
%   \centering
%   \includegraphics[width=\linewidth]{samples/imgs/cross.pdf}
%   \vspace{-12pt}
%   \caption{The cross-model annotation-validation process. }
%   \label{cross}
% \end{figure}
\subsection{Towards Explainable Quality Assessment}

The ultimate objective of EPAIQA is to deliver a \textbf{reasonable and explainable} feedback of the  editing quality of PAIs. So in this stage, we primarily emphasize the CoT data injection for further SFT, as outlined on the right side of Fig. \ref{model}. Additionally, substituting the costly manual check,  we implement the \textbf{cross-model annotation-validation process} for data validation (shown in Fig. \ref{cross}).

First, we select two representative quality scenarios in the \textbf{overall quality subset}  as the basis for CoT data  generation: (1) scenario where the overall editing quality is poor due to low prompt completion (level $1$ and level $2$); (2) scenario where the prompt completion is level $3$. Next, for each sample, we randomly select one model (the \textbf{annotator}) from three state-of-the-art proprietary LMMs (\textit{Gemini-1.5-Pro}, \textit{GPT-4o}, and \textit{Claude-3.7-Sonnet} \cite{anthropic2025claude}) to generate explanatory CoT pieces for the two above-mentioned scenarios. For the former, consistent with the subjective experiment setup, the annotator is only required to provide an analytical description indicating why the prompt is not followed in the given editing instance. For the latter, the model provides descriptions for prompt completion, naturalness, and harmony—based on the given human-labeled quality level. In both cases, the model's outputs are used as CoT pieces for subsequent merging. For each generated CoT piece, we use the other two models (\textbf{scrutinizers}) to validate the content. Only when both scrutinizers agree that the CoT explanation is reasonable will it be deemed qualified.

The rationale behind this design is that we believe that prompt completion assessment is one of the primary tasks for the model to interpret the editing quality. Meanwhile, the quality analysis of naturalness and harmony represents the understanding of the local consistency of the edited region and the overall coherence of the PAI. A step-by-step analysis of these three components \textbf{aligns with the cognitive process typically employed by humans} when assessing the quality of a PAI.
 Furthermore, by performing CoT injection on different scenarios, which include a variety of quality levels across the dimensions, we ensure a balance between positive and negative examples. This effectively prevents the model’s reasoning process from overfitting to a particular scenario, ensuring the reasoning process's generalization ability. The cross-validation process effectively guarantees the quality of annotations without human-in-the-loop and enhances the diversity of annotation styles, facilitated by the random model selection.

After all these processes, we perform merging and refinement to obtain an effective attribution analysis for the overall quality level (as shown in the lower-right part of Fig. \ref{model}) and obtain the stage-3 instruction dataset. Consequently, the stage-3 model is built upon the stage-2 model via training on this dataset.

\vspace{-2pt}
\section{Experiments}
\subsection{Experiments Setups}
\label{Experiments Setups}
\noindent \textbf{Overall setups} We use \textit{llava-onevision-qwen2 (siglip) (\textit{7}B)} \cite{llava} as the base model for training. In the stage-1 training process, we freeze the LLM component, whereas in the subsequent stage-2 and stage-3, we perform full-parameter SFT. For image pre-processing, we adopt the \textbf{Anyres} mode to ensure that input images are loaded approximately at their original resolution. 

\begin{table*}[t]\tiny
    \centering
    \renewcommand\arraystretch{1.15}
    \renewcommand\tabcolsep{7.5pt}
    \belowrulesep=0pt\aboverulesep=0pt
    
    \caption{Performance on global harmony scoring task. Type 1-4 denote style change, object enhancement,  object operation, and
      semantic change, respectively. [\textcolor{red}{red}: the best result, \textcolor{blue}{blue}: the second best result]}
\vspace{-10pt}
    \resizebox{1\linewidth}{!}
    {\begin{tabular}{l|cc|cc|cc|cc|cc}
 \hline
    \multicolumn{1}{l|}{\textbf{Manipulation Type}}&\multicolumn{2}{c|}{\textbf{Type-1}}& 
    \multicolumn{2}{c|}{\textbf{Type-2} }&\multicolumn{2}{c|}{\textbf{Type-3} }&\multicolumn{2}{c|}{\textbf{Type-4 }}&\multicolumn{2}{c}{\textbf{Overall} }\\ 
    \cdashline{1-11}
       \multicolumn{1}{l|}{\textit{\textbf{Models}}}&SRCC &PLCC &SRCC &PLCC &SRCC &PLCC &SRCC &PLCC &SRCC &PLCC \\ \cdashline{1-11}

      \textit{DBCNN (TCSVT 2020)} \cite{DBCNN}  & 0.301& 0.272 & 0.402 & 0.355 & 0.185& 0.214 & 0.372& 0.388&0.353 &0.325  \\
      \textit{Hyper-IQA (CVPR 2020)} & 0.342 & 0.322 & 0.431& 0.406 & 0.308& 0.301& 0.398& 0.401& 0.402& 0.387\\
     \textit{MUSIQ (CVPR 2021)} \cite{MUSIQ} & 0.321& 0.337 & 0.407& 0.399 & 0.368&0.337 &0.363 &0.382 &0.350 &0.332  \\
      \textit{QualiCLIP (Arxiv 2025)} \cite{QualiCLIP} & 0.346& 0.327 & 0.482& 0.432 &0.418&0.404&0.522 &0.495 & 0.400&  0.384 \\
       \textit{Stair-IQA (JSTSP 2023)} \cite{Stair-IQA} &0.289 & 0.301 & 0.351&  0.362&0.320 & 0.333& 0.348& 0.384& 0.342&  0.355 \\
       \textit{NIMA (TIP 2018)} \cite{NIMA} &0.304 &0.269  &0.448 &0.318  &0.361 &0.310 &0.411 &0.400 &0.378 &0.343 \\
        \textit{TOPIQ-NR (TIP 2024)} \cite{TOPIQ} &0.416 &0.410  &0.443 &0.408  &0.250 &0.223 &0.418 &0.425 &0.376 & 0.351 \\
        \textit{WaDIQaM-NR} \cite{WaDIQaM}  &0.061 &0.052  &0.204 &0.287  &0.119 &0.155 &0.111 &0.160 &0.156 &0.137  \\
         \textit{TOPIQ-FR} &0.356 &0.397  & 0.371&0.466  & 0.411& {\color{blue}0.472}&0.519 &{\color{blue}0.538} & 0.444& 0.463\\
          \textit{WaDIQaM-FR}  & 0.237& 0.259 &0.251 &  0.257&0.208 &0.221 & 0.254& 0.272& 0.265& 0.258  \\

       \textit{q-align-IQA (\textit{7B})} \cite{qalign} &0.543 &0.521  &0.512 &0.493  &0.445 &0.413 &0.519 &0.502 &0.473 &0.463  \\ 
    \textit{q-align-onealign (\textit{7B})} &{\color{blue}0.557} &{\color{blue}0.537}  &{\color{blue}0.520} &{\color{blue}0.503}  &{\color{blue}0.453} &0.420 &{\color{blue}0.531} &0.522 &{\color{blue}0.505} &{\color{blue}0.507} \\ 
   \cdashline{1-11}
     \textit{EPAIQA-Stage2 (\textit{7B})}   &{\color{red}0.718} &{\color{red}0.750}  &{\color{red}0.564} &{\color{red}0.667} &{\color{red}0.695} &{\color{red}0.738} &{\color{red}0.716} &{\color{red}0.723} &{\color{red}0.703} &{\color{red}0.740}   \\
     % &\textit{stage3}& & 0.7418~/~0.7855 & 0.8683~/~0.8358   &   \\
     % &\textit{VQA\textsuperscript{2}-stage2}& 0.8253~/~ 0.8568 &0.7513~/~0.8117  &0.8756~/~0.8649 & 0.8843~/~0.8798     \\
     % &\textit{VQA\textsuperscript{2}-scorer-7B (mix)}&0.8266~/~0.8675 & 0.7609~/~0.8308 & 0.8825~/~0.8733 & 0.895~/~0.892 \\
  \hline
    \end{tabular}}
    \vspace{-8pt}

    \label{tab:rating1}
\end{table*}
\begin{table*}[t]\tiny
    \centering
    \renewcommand\arraystretch{1.15}
    \renewcommand\tabcolsep{7.5pt}
    \belowrulesep=0pt\aboverulesep=0pt

    \caption{Performance on local naturalness scoring task. }
    \vspace{-11pt}
    \resizebox{1\linewidth}{!}
    {\begin{tabular}{l|cc|cc|cc|cc|cc}
 \hline
    \multicolumn{1}{l|}{\textbf{Manipulation Type}}&\multicolumn{2}{c|}{\textbf{Type-1}}& 
    \multicolumn{2}{c|}{\textbf{Type-2} }&\multicolumn{2}{c|}{\textbf{Type-3} }&\multicolumn{2}{c|}{\textbf{Type-4}}&\multicolumn{2}{c}{\textbf{Overall} }\\ 
    \cdashline{1-11}
       \multicolumn{1}{l|}{\textit{\textbf{Models}}}&SRCC &PLCC &SRCC &PLCC &SRCC &PLCC &SRCC &PLCC &SRCC &PLCC \\ \cdashline{1-11}

      \textit{DBCNN} &0.378 &0.378 & 0.446 &0.412 &0.274 & 0.239& 0.383& 0.353& 0.339& 0.358\\
      \textit{Hyper-IQA} \cite{HyperIQA}  & 0.382 & 0.353 & 0.411& 0.418 & 0.326& 0.319& 0.401& 0.408& 0.428& 0.417\\
     \textit{MUSIQ} &0.294& 0.298 & 0.354& 0.332 & 0.307&0.314 &0.344 &0.339 &0.336 &0.321\\
      \textit{QualiCLIP } &0.411 &0.401  & 0.471& 0.447 &0.344 &0.348 & 0.437 & 0.424 & 0.379& 0.393  \\
       \textit{Stair-IQA } & 0.369&  0.380& 0.399& 0.420 & 0.363& 0.362& 0.357&0.328 & 0.387&  0.365 \\
       \textit{NIMA}  &0.447 &0.411  &0.409 &0.374  &0.427 &0.403 &0.390 &0.358 &0.364 &0.387 \\
        \textit{TOPIQ-NR} &0.416 &0.410  &0.443 &0.408  &0.250 &0.223 &0.418 &0.425& 0.407&0.384  \\
        \textit{WaDIQaM-NR (CVPR 2021)}   &0.055 &0.149  &0.164 &0.148  &0.186 &0.044 &0.115 &0.190 &0.188 &0.193\\
         \textit{TOPIQ-FR} &0.388 & 0.404 & 0.351&0.418  & 0.202& 0.212& 0.415&0.434 & 0.410 & 0.383 \\
          \textit{WaDIQaM-FR} &0.293 & 0.298 &0.306 & 0.307 &0.293 & 0.297&0.271 & 0.274& 0.268& 0.263\\

           \textit{JOINT (TCSVT 2024)} \cite{JOINT}&{\color{blue}0.602} &{\color{blue}0.598} &{\color{blue}0.565} &0.517 &{\color{blue}0.493}&{\color{blue}0.448} &{\color{blue}0.532} &{\color{blue}0.565} &{\color{blue}0.613} &{\color{blue}0.635}   \\
       \textit{q-align-IQA (\textit{7B})} &0.512 &0.483 &{0.533} &0.521 &0.452 &0.423 &0.448 &0.465 &0.482 &0.477  \\ 
    \textit{q-align-onealign (\textit{7B})}&0.551 &0.495 &0.517 &{\color{blue}0.530} &0.470 &0.445 &0.460 &0.480 &0.503 &0.485  \\ 
   \cdashline{1-11}
   \textit{EPAIQA-Stage2 (\textit{7B})}&{\color{red}0.703} &{\color{red}0.725} &{\color{red}0.715} &{\color{red}0.708} &{\color{red}0.576} &{\color{red}0.603} &{\color{red}0.667} &{\color{red}0.708} &{\color{red}0.674} &{\color{red}0.702}   \\

     % \textit{EPAIQA-Stage2 (\textit{7B})}&{\color{red}0.703} &0.725 &0.715 &0.708 &0.576 &0.603 &0.667 &0.708 &0.674 &0.702   \\
     % &\textit{stage3}& & 0.7418~/~0.7855 & 0.8683~/~0.8358   &   \\
     % &\textit{VQA\textsuperscript{2}-stage2}& 0.8253~/~ 0.8568 &0.7513~/~0.8117  &0.8756~/~0.8649 & 0.8843~/~0.8798     \\
     % &\textit{VQA\textsuperscript{2}-scorer-7B (mix)}&0.8266~/~0.8675 & 0.7609~/~0.8308 & 0.8825~/~0.8733 & 0.895~/~0.892 \\
  \hline
    \end{tabular}}
      \vspace{-10pt}
    \label{tab:rating2}
\end{table*}
% For the stage-2 and stage-3 training and test sets, we utilize non-overlapping datasets partitioned according to an 80\%-20\% split, as described previously.
\noindent \textbf{Setups for stage-2 model evaluation}
We first employ the stage-2 model to perform quantitative scoring of the edited images' local naturalness and global harmony. As these two dimensions in the subjective experiments are rated solely based on the edited images, the experiment permits both single-image and full-referenced input. Consequently, we select state-of-the-art FR and NR methods from the IQA domain as comparing methods, including the LMM-based \textit{q-align} \cite{qalign} and the \textit{JOINT} \cite{JOINT} designed for AGI naturalness assessment. We select the naturalness subset and harmony subset mentioned in Sec. \ref{split} for training and testing for comparing models (except for \textit{q-align}). In here and subsequent scoring testing experiments, we adopt the scoring paradigm from \textit{q-align}, which involves applying softmax to the output logits of the quality level keywords and performing a weighted summation. We adopt the commonly used \textit{Pearson Linear Correlation Coefficient} (PLCC) and  \textit{Spearman Rank Correlation Coefficient} (SRCC) as the evaluation metrics, with the experiments' results on overall harmony and local naturalness presented in Tabs. \ref{tab:rating1} and \ref{tab:rating2}, respectively.

\noindent \textbf{Setups for stage-3 model evaluation}
For stage-3 model evaluation, we select the samples in the test set of \textbf{overall quality subset} (designed in Sec. \ref{split}) with a prompt completion of $3$ for model testing ($682$ samples). This selection criterion is based on the fact that traditional IQA models (the comparing models) are generally incapable of assessing high-level factors, such as prompt completion. Furthermore, due to the intricate input requirements (typically two images and an editing prompt), there are currently no comparable methods within the specific domain that can accommodate such inputs. As a result, we devise a specialized approach for scoring comparison. For the FR and NR IQA model types, we initially utilize the models that demonstrate the highest performance in predicting naturalness (FR: \textit{TOPIQ-FR}, NR: \textit{JOINT}) and harmony (FR: \textit{TOPIQ-FR}, NR: \textit{q-align}) scores in previous experiments on the training set to output the corresponding sub-dimension scores. Following this, we employ \textit{random forest regression} \cite{developers2019scikit} to derive the optimal regressors that map the two sub-dimension scores to the overall quality score. The optimized NR and FR regressors are referred to as the `\textbf{best competitor regressors}' and are used for comparison experiments, with the results presented in Tab. \ref{tab:rating3}.

Furthermore, to assess the explainable quality evaluation capability of the stage-3 model, we select and manually review and refine $300$ samples from the test set, which are annotated with CoT (including various scenarios across different quality dimensions) as the test set for this task
. We employ the \textit{o3-mini} model to gauge (using a voting strategy similar to \cite{zhang2024q}) the consistency of models' responses with the answers in the test set in four evaluation concerns: the prompt completion judgment accuracy (\textbf{PA}), the local naturalness analysis alignment (\textbf{LNA}), the
 global harmony analysis alignment  (\textbf{GHA}), and the overall reasoning alignment (\textbf{Overall}). The judgment criteria, the voting process, and more details about this experiment are detailed in \textit{supp.} Sec. \ref{Evaluation}.  We select several open-source LMMs with robust high-level image understanding and analysis capabilities as comparison models, with the specific results presented in Tab. \ref{tab:rating4}.
 \begin{figure*}[t]
  \label{11}
  \centering
  \includegraphics[width=0.99\linewidth]{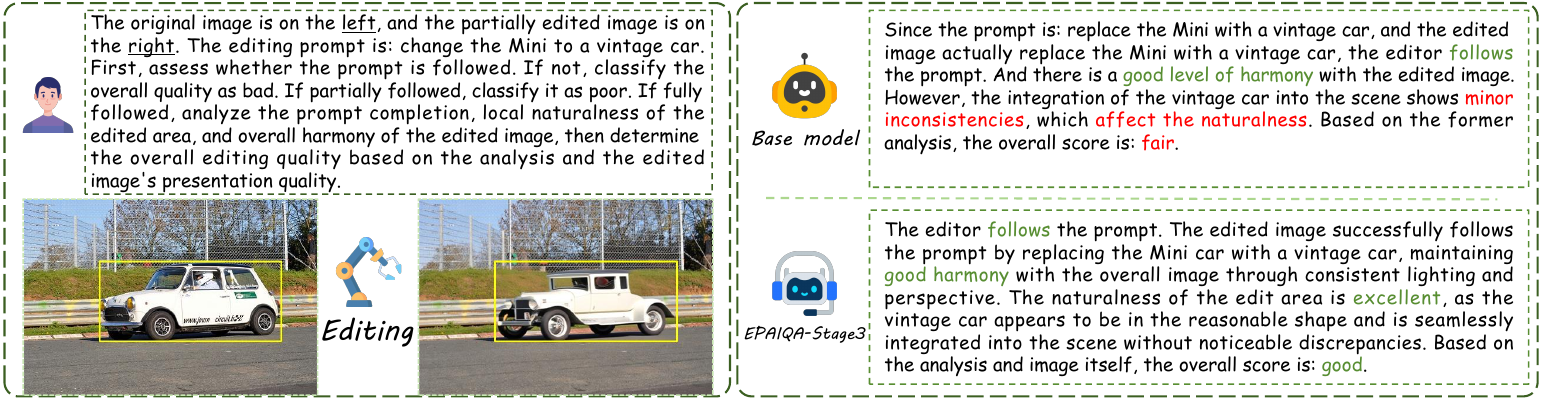}
  \vspace{-8pt}
  \caption{ Case study examples of stage-3 model performance. [\textcolor{green}{green}: the correct judgment, \textcolor{red}{red}: the incorrect judgment]]}
  \label{case study1}
\end{figure*}
\begin{table*}[t]\tiny
    \centering
    \renewcommand\arraystretch{1.15}
    \renewcommand\tabcolsep{8.2pt}
    \belowrulesep=0pt\aboverulesep=0pt

    \caption{Performance on overall editing quality scoring task. [\textcolor{red}{red}: the best result.]}
    \vspace{-10pt}
    \resizebox{1\linewidth}{!}
    {\begin{tabular}{l|cc|cc|cc|cc|cc}
 \hline
    \multicolumn{1}{l|}{\textbf{Manipulation Type}}&\multicolumn{2}{c|}{\textbf{Type-1}}& 
    \multicolumn{2}{c|}{\textbf{Type-2} }&\multicolumn{2}{c|}{\textbf{Type-3} }&\multicolumn{2}{c|}{\textbf{Type-4 }}&\multicolumn{2}{c}{\textbf{Overall} }\\ 
    \cdashline{1-11}
       \multicolumn{1}{l|}{\textit{\textbf{Metrics}}}&SRCC &PLCC &SRCC &PLCC &SRCC &PLCC &SRCC &PLCC &SRCC &PLCC \\ \cdashline{1-11}
          \textit{Best competitor (NR-IQA)} & 0.387 & 0.392&0.451&0.462&0.401&0.421&0.423&0.429&0.417&0.409\\
      \textit{Best competitor (FR-IQA)} & 0.412 & 0.439&0.449&0.459&0.438&0.441&0.451&0.449&0.462&0.466\\
      
       \cdashline{1-11}
     \textit{EPAIQA-Stage3 
 % (\textit{7B})}&0.572 &0.589  &0.716 &0.735  &0.525 &0.553  &0.615 &0.622 &0.631 &0.659  \\
 (\textit{7B})}&{\color{red}0.572} &{\color{red}0.589}  &{\color{red}0.716} &{\color{red}0.735}  &{\color{red}0.525} &{\color{red}0.553}  &{\color{red}0.615} &{\color{red}0.622} &{\color{red}0.631} &{\color{red}0.659}  \\
   
  \hline
    \end{tabular}}
      \vspace{-12pt}
    \label{tab:rating3}
\end{table*}

\begin{table}[t]\tiny
    \centering
    \renewcommand\arraystretch{1.05}
    \renewcommand\tabcolsep{5.5pt}
    \belowrulesep=0pt\aboverulesep=0pt

    \caption{Performance on quality explanation tasks.}
    \vspace{-10pt}
    \resizebox{1\linewidth}{!}
    {\begin{tabular}{l|c|c|c|c}
 \hline
    \multicolumn{1}{l|}{\textbf{LMMs}}&\multicolumn{1}{c|}{\textbf{PA}}& 
    \multicolumn{1}{c|}{\textbf{LND} }&\multicolumn{1}{c|}{\textbf{GHD} }&\multicolumn{1}{c}{\textbf{Overall}}\\ 
    \cdashline{1-5}
       
       \textit{LLaVA-ov(7B)  (base model)} \cite{llava}&0.368 &0.432 &0.453 &0.351 \\
      \textit{QwenVL-2 (7B)} \cite{QwenVL-2}&0.452 &0.507 &0.482 &0.423 \\
      \textit{QwenVL-2.5 (7B)} \cite{QwenVL-2.5}&0.431 &0.492 &0.467 &0.415  \\
      \textit{DeepseekVL-2 (7B)} \cite{wu2024deepseek} &0.393 &0.423 &0.446  &0.385  \\
      \textit{Q-instruct (LLaVA) (7B)} \cite{Q-instruct}&0.307 &0.505 &0.511 &0.315 \\
       \textit{Aes-expert (7B)} \cite{Aes-expert}&0.355 &0.527 &0.476 &0.343 \\
       \cdashline{1-5}
     \textit{EPAIQA-Stage3 
 % (\textit{7B})} &0.683&0.585&0.558&0.661\\
   (\textit{7B})} &{\color{red}0.683}&{\color{red}0.585}&{\color{red}0.558}&{\color{red}0.661}\\
  \hline
    \end{tabular}}
      \vspace{-11pt}
    \label{tab:rating4}
\end{table}
\begin{table}[t]\small
    \centering
    \renewcommand\arraystretch{1.12}
    \renewcommand\tabcolsep{3pt}
    \belowrulesep=0pt\aboverulesep=0pt
   
    \caption{Comparison of the \textit{overall quality scoring} performance (SRCC$\uparrow$/PLCC$\uparrow$) with/without CoT injection.}
     \vspace{-10pt}
    \resizebox{1\linewidth}{!}
    {\begin{tabular}{c|ccccc}
    \hline
     \multicolumn{1}{c|}{\textbf{Version}}&{\textit{Type-1}} & {\textit{Type-2}} & {\textit{Type-3}} & \textit{Type-4} & \textit{Overall}\\ \hline 
     \textit{wo CoT}&{\color{red}0.575}~/~0.582 & 0.711~/~0.727 & 0.518~/~0.544 & 0.608~/~0.617 &0.623~/~0.650\\
     % &\textit{VQA\textsuperscript{2}-scorer-7B (mix)}&0.8266~/~0.8675 & 0.7609~/~0.8308 & 0.w8825~/~0.8733 & 0.895~/~0.892 \\
      \textit{w CoT}&0.572~/~{\color{red}0.589}  &{\color{red}0.716}~/~{\color{red}0.735}  &{\color{red}0.525}~/~{\color{red}0.553}  &{\color{red}0.615} ~/~{\color{red}0.622} &{\color{red}0.631}~/~{\color{red}0.659} \\
    \hline
    \end{tabular}}
    
    \label{tab:cot}
\end{table}
\vspace{-6pt}
\subsection{Experiments Results}
\noindent \textbf{Results for stage-2 model evaluation}
We have the following observations. First, our model shows significant improvements over most comparison models across almost all editing types, which validates the rationality of our model design. Moreover, we observe that the approach utilizing the LMM as the base model (such as \textit{q-align}), even without additional training, surpasses the performance of most traditional models trained on our dataset. This further demonstrates the superiority of LMMs in performing PAIQA naturalness and harmony assessments.

\noindent \textbf{Results for stage-3 model evaluation}
For the overall quality scoring task,  we observe that our model outperforms the competitors with a significant margin, while the performance of the best competitors is generally sub-optimal. We attribute this to their intrinsic limitations in accurately predicting naturalness and harmony levels, as well as the inability of simple regression methods (such as SVR and RF) to effectively capture the complex relationship between naturalness, harmony, and overall quality.

For the quality explanation task, after training with the stage-3 dataset incorporating CoT, our model exhibits significantly enhanced quality feedback capabilities compared to other LMMs with comparable size, including both general LMMs \cite{llava,QwenVL-2,QwenVL-2.5,wu2024deepseek} and those designed for visual quality assessment \cite{Q-instruct,Aes-expert}. This demonstrates that the inclusion of CoT significantly enhances the interpretability of the quality assessment LMM, providing useful EPAIQA feedback. Furthermore, to visualize this capability, we provide several test \textbf{case studies}  in Fig. \ref{case study1} and \textit{supp.} Sec. \ref{case}.
\vspace{-6pt}
\subsection{Ablation Study}
\noindent \textbf{The effects of the use of CoT } The injection of CoT primarily enhances the model's ability to provide \textbf{explainable quality feedback} and overall quality level \textbf{reasoning}. It is the capability that is absent in conventional scoring-centric models. For further exploration, we train the stage-2 model with two separate instruction sets: one incorporating CoT and another directly using the overall editing quality level as the training target. We then compare the two models' scoring ability under the same experimental conditions as mentioned before, with the results shown in Tab. \ref{tab:cot}. The model incorporating CoT data also demonstrates a noticeable performance improvement in the scoring task. Thus, we posit that the key role of CoT injection lies in enabling explainable feedback, and its quality reasoning capacity also provides site-effect to improve the performance of quantitative scoring.

\noindent \textbf{The effects of grounding pre-training}
The primary goal of grounding pre-training is to enhance the model's sensitivity to AI-generated features within the edited regions, thereby enhancing its performance in scoring tasks requiring whole-image input—such as harmony level prediction and overall quality assessment—when bbox information is unavailable in the evaluated images. To further explore this effect, we retrain the stage-2 and stage-3 models using two alternative approaches. First, we omit stage-1 and directly train the stage-2 and stage-3 models, referred to as the \textbf{`wo-stage-1'} training mode. For comparison, we also exclude stage-1 pre-training but use edited images with bboxes for training the stage-2 and stage-3 models (with the training and testing instruction specifying the bbox area as the edited region). This is referred to as the \textbf{`reference'} mode. The ablation comparison results are shown in Tabs. \ref{tab:harmony_grounding} and \ref{tab:overall_grounding}. The experimental results indicate that under the `wo-stage-1' training mode, both models exhibit a notable decline in scoring performance. When compared with the `reference', the model with stage-1 shows only a slight performance difference. This demonstrates the impact of the grounding pre-training process.

\begin{table}[t]\small
    \centering
    \renewcommand\arraystretch{1.15}
    \renewcommand\tabcolsep{2.7pt}
    \belowrulesep=0pt\aboverulesep=0pt
   
    \caption{Ablation study of the \textit{stage-2 harmony} scoring performance (SRCC$\uparrow$/PLCC$\uparrow$) for grounding pre-training. [\textcolor{red}{red}: the best result]}
     \vspace{-9pt}
    \resizebox{1\linewidth}{!}
    {\begin{tabular}{c|ccccc}
    \hline
     \multicolumn{1}{c|}{\textbf{Version}}&{\textit{Type-1}} & {\textit{Type-2}} & {\textit{Type-3}} & \textit{Type-4} & \textit{Overall}\\ \hline 
     \textit{wo stage-1}&0.675~/~0.733 & 0.485~/~0.613 &0.667~/~0.725  &0.698~/~0.702&0.667~/~0.708 \\
     \textit{reference}&{\color{red}0.735}~/~0.741 & {\color{red}0.580}~/~{\color{red}0.673} & {\color{red}0.707}~/~0.729 & {\color{red}0.720}~/~{\color{red}0.725} & {\color{red}0.710}~/~{\color{red}0.745} \\
     % &\textit{VQA\textsuperscript{2}-scorer-7B (mix)}&0.8266~/~0.8675 & 0.7609~/~0.8308 & 0.w8825~/~0.8733 & 0.895~/~0.892 \\
      \textit{w stage-1}&0.718~/~{\color{red}0.750}  &0.564~/~0.667 &0.695~/~{\color{red}0.738} &0.716~/~0.723 &0.703~/~0.740 \\
    \hline
    \end{tabular}}
    \vspace{-12pt}
    \label{tab:harmony_grounding}
\end{table}

\begin{table}[t]\small
    \centering
    \renewcommand\arraystretch{1.15}
    \renewcommand\tabcolsep{2.7pt}
    \belowrulesep=0pt\aboverulesep=0pt
   
    \caption{Ablation study of the \textit{stage-3 overall quality scoring} performance (SRCC$\uparrow$/PLCC$\uparrow$) for grounding pre-training.}
     \vspace{-10pt}
    \resizebox{1\linewidth}{!}
    {\begin{tabular}{c|ccccc}
    \hline
     \multicolumn{1}{c|}{\textbf{Version}}&{\textit{Type-1}} & {\textit{Type-2}} & {\textit{Type-3}} & \textit{Type-4} & \textit{Overall}\\ \hline 
     \textit{wo stage-1}&0.542~/~0.567 & 0.707~/~0.716 &0.519~/~0.550 & 0.599~/~0.603  &0.614~/~0.635\\
     \textit{reference}&{\color{red}0.577}~/~{\color{red}0.593} & {\color{red}0.723}~/~0.721 & {\color{red}0.533}~/~{\color{red}0.556} & {\color{red}0.620}~/~{\color{red}0.625} &{\color{red}0.640}~/~{\color{red}0.661} \\
     % &\textit{VQA\textsuperscript{2}-scorer-7B (mix)}&0.8266~/~0.8675 & 0.7609~/~0.8308 & 0.w8825~/~0.8733 & 0.895~/~0.892 \\
      \textit{w stage-1}&0.572~/~0.589  &0.716~/~{\color{red}0.735}  &0.525~/~0.553  &0.615 ~/~0.622 &0.631~/~0.659 \\
    \hline
    \end{tabular}}
    \vspace{-10pt}
    \label{tab:overall_grounding}
\end{table}

\section{Conclusion}

We propose the \textbf{EPAIQA-$15$K dataset}, the first large-scale dataset for PAIQA with over $15,000$ edited images, and develop the \textbf{EPAIQA series models} with interpretable quality feedback capabilities. The model construction pipeline is divided into three stages, comprehensively covering the quality scoring and quality explanation tasks with abundant CoT data. The EPAIQA series models perform excellently in the PAIQA scoring tasks, significantly surpassing existing IQA models. Furthermore, the model provides the CoT feedback capability, effectively enhancing the model's versatility. Our work offers compelling insights for the PAIQA-related field, laying a solid foundation for subsequent research.
{

\bibliographystyle{ACM-Reference-Format}
\bibliography{acmart}}

\clearpage
\appendix

\begin{figure*}[htbp]
    \centering
    \begin{subfigure}[b]{0.48\linewidth}    
        \centering
        \includegraphics[width=0.95\linewidth]{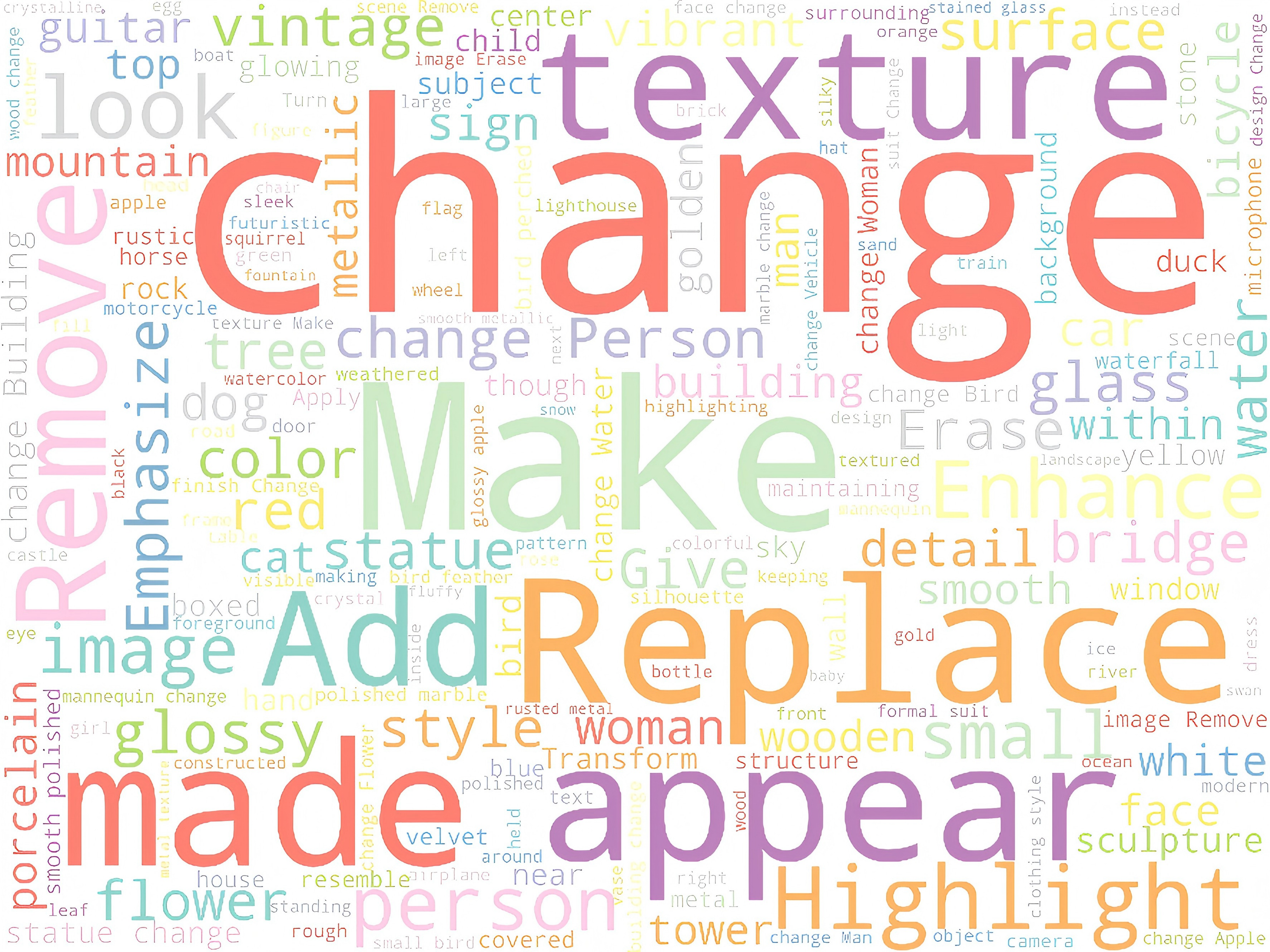}
        \caption{Editing prompts}
        \label{fig:cloud1}
    \end{subfigure}
    \begin{subfigure}[b]{0.48\linewidth}
        \centering
        \includegraphics[width=0.95\linewidth]{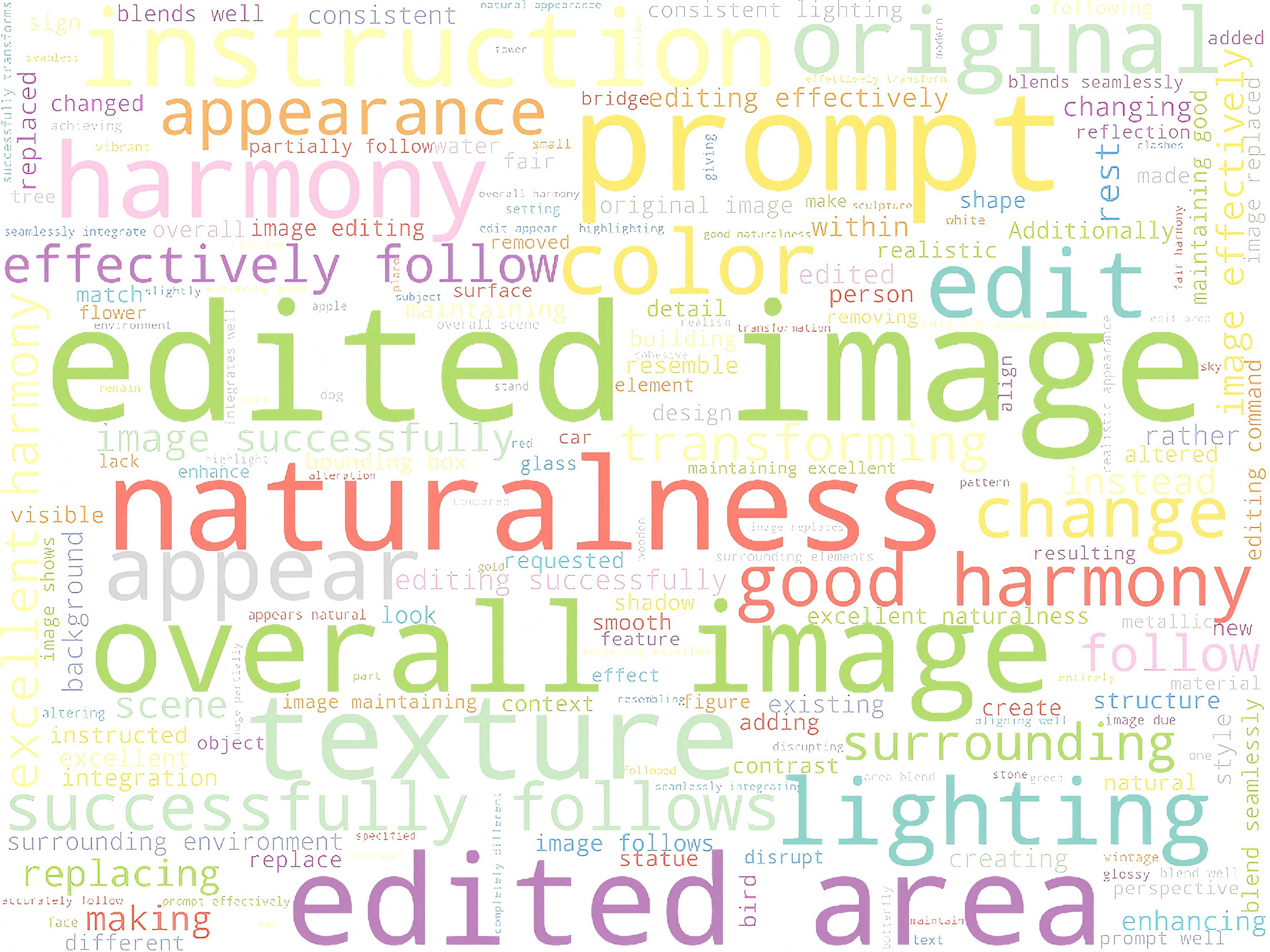}
        \caption{CoTs}
        \label{fig:cloud2}
    \end{subfigure}
    \vspace{-12pt}
    \caption{Word clouds}
    \label{fig:cloud}
\end{figure*}

\begin{table*}[htbp]\small
\caption{The descriptions of three dimensions}
\vspace{-5pt}
    \centering
    \begin{tabular}{p{4cm}|p{12cm}}
        \hline
        \textbf{Dimension} & \textbf{Description} \\
        \hline
        \textbf{Prompt Completion } & The edited area is evaluated to determine whether it adheres to the requirements outlined in the prompts \\
        \hdashline
        \textbf{Harmony} & The edited area is evaluated based on whether it aligns with the overall spatial logic (such as object placement, local and global semantic relationships). Additionally, the consistency of details such as color and lighting with the surrounding areas is assessed. The edited area is also assessed to determine whether there are any noticeable traces of editing when compared to the surrounding regions. \\
        \hdashline
        \textbf{Naturalness of the edited area}  & The edited area is assessed to determine whether it appears natural. \\
        \hline
    \end{tabular}
    
    \label{tab:threedimension}
\end{table*}

\section{Key Patterns Visualization}

We generate separate \textbf{word clouds} for the editing prompts and the CoTs to provide a visual representation of their respective key terms in Fig. \ref{fig:cloud}
% \begin{figure*}[htbp]
%     \centering
%     \begin{subfigure}[b]{0.48\linewidth}    
%         \centering
%         \includegraphics[width=0.95\linewidth]{samples/imgs/Prompt1.pdf}
%         \caption{Editing prompts}
%         \label{fig:cloud1}
%     \end{subfigure}
%     \begin{subfigure}[b]{0.48\linewidth}
%         \centering
%         \includegraphics[width=0.95\linewidth]{samples/imgs/CoT1.pdf}
%         \caption{CoTs}
%         \label{fig:cloud2}
%     \end{subfigure}
%     \vspace{-12pt}
%     \caption{Word clouds}
%     \label{fig:cloud}
% \end{figure*}
\section{Supplementary Details for EPAIQA-15K}
\label{APED:example}
\begin{figure}[htbp]
  \centering
  \includegraphics[width=0.9\linewidth]{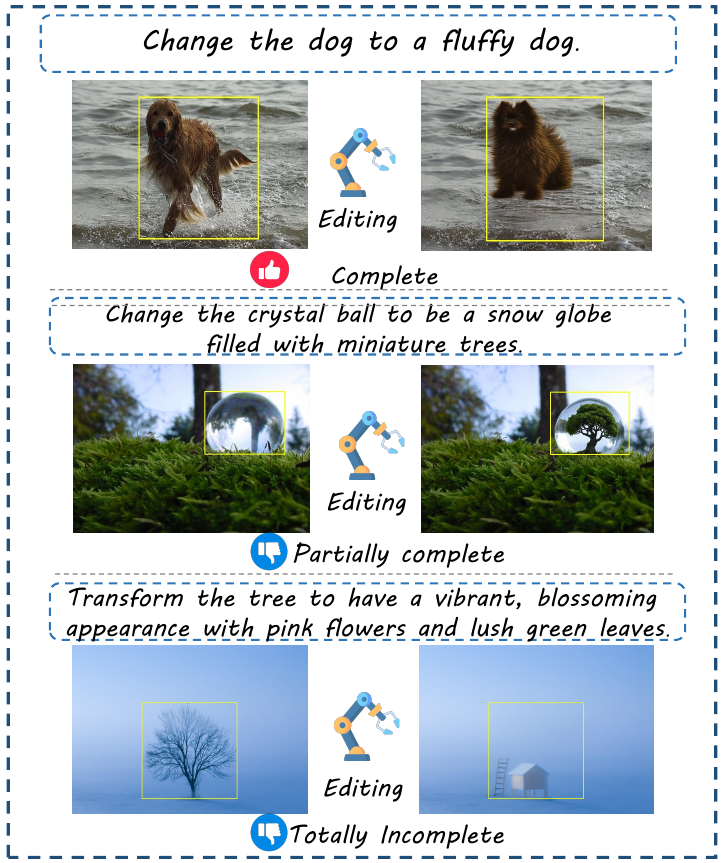}
  \vspace{-8pt}
  \caption{Examples of different prompt completion level.}
   \label{pc}
\end{figure}

\begin{figure*}[htbp]
  \centering
  \includegraphics[width=0.85\linewidth]{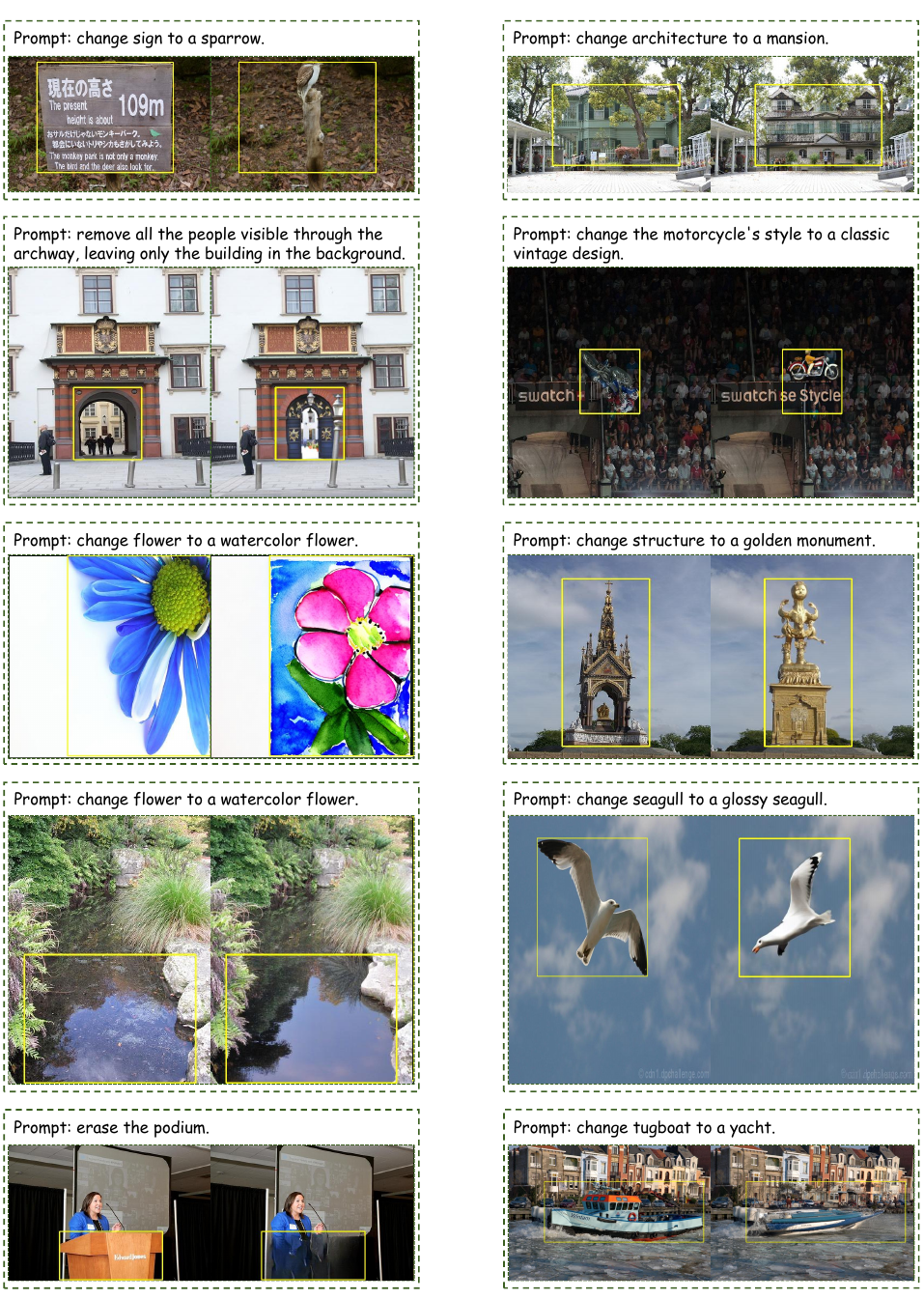}
  \vspace{-8pt}
  \caption{Illustrative examples of data pairs demonstrating strict adherence to the provided prompts.}
   \label{appendixA}
\end{figure*}

Our \textbf{EPAIQA-15K dataset} consists of original images, edited images, and corresponding prompts.Each data pair is rigorously curated, with prompts accurately describing the edited subject to ensure precise text-visual correspondence. All images comply with rigorous quality standards, validated through a combined manual and automated approach to ensure the dataset's reliability. We randomly select a few examples that strictly follow the prompts for display, as shown in the Fig. \ref{appendixA}. We also provide examples of different prompt completion level in the Fig. \ref{pc}

% \begin{figure*}[htbp]
%   \centering
%   \includegraphics[width=0.85\linewidth]{samples/imgs/appendixA.pdf}
%   \vspace{-8pt}
%   \caption{Illustrative examples of data pairs demonstrating strict adherence to the provided prompts.}
%    \label{appendixA}
% \end{figure*}

\section{Subjective Experiment Settings}
\label{experiment}
\subsection{Overview of Experiment Settings}
\begin{figure*}[htbp]
  \centering
  \includegraphics[width=0.95\linewidth]{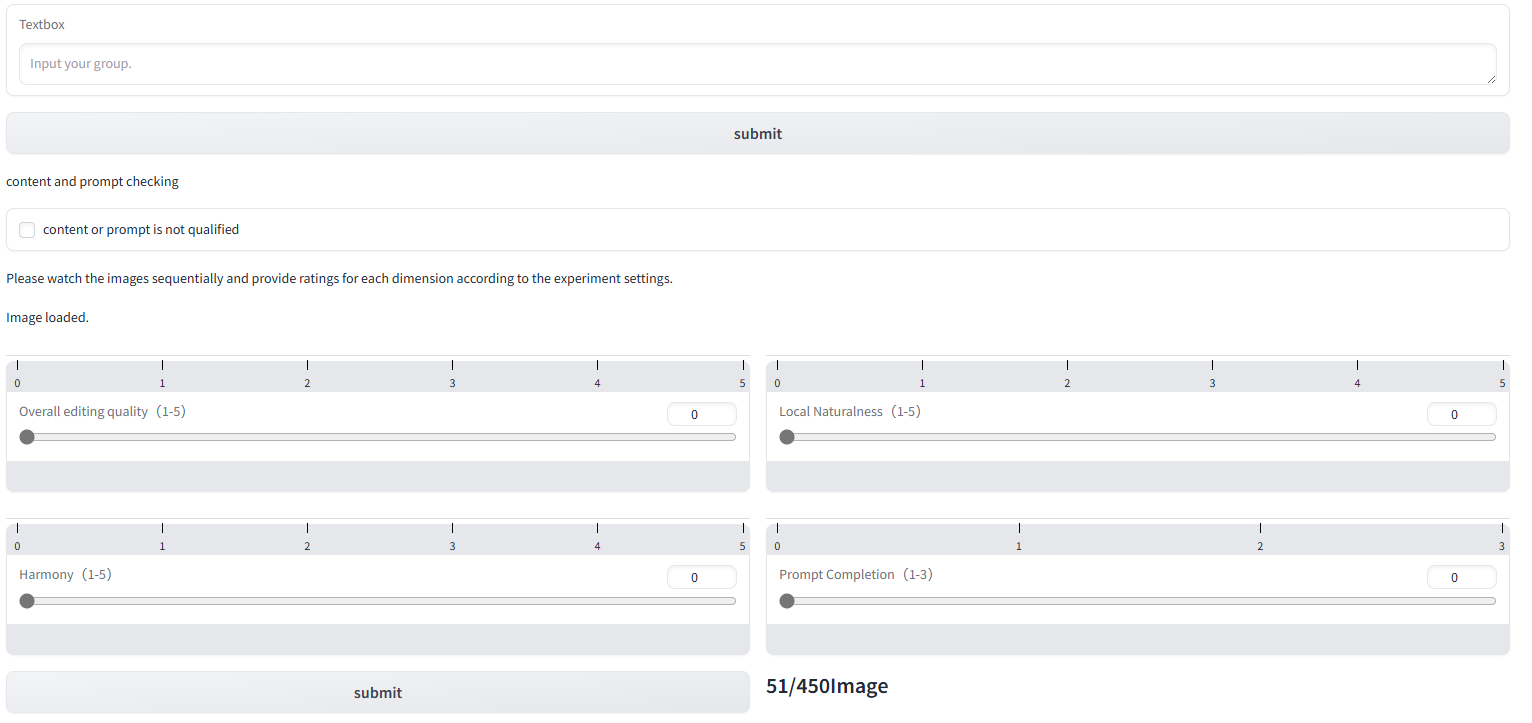}
  \vspace{-8pt}
  \caption{The interface screenshot used for participant scoring.}
   \label{screenshot}
\end{figure*}

The participants are required to view the original image (left), the edited image (middle), and the edited image with a bounding box (right), along with the corresponding prompt. They rate the image-prompt pairs on \textbf{three dimensions} and provide an \textbf{overall score for the editing quality}. The area outlined by the bounding box is the edited region. The explanations of the three dimensions are shown in Tab. \ref{tab:threedimension}. The interface screenshot is shown in Fig. \ref{screenshot}.

\subsection{Prompt and Data Check}
Before proceeding with scoring, it is essential to verify whether the prompt meets the specified requirements and whether the image content complies with the established criteria. Instances of prompts or images that meet the standards below should be annotated and excluded from the scoring process.
\begin{itemize}
    \item \textbf{The subject of editing specified in the prompt is entirely unrelated to the primary object within the outlined editing region.}
    \item The prompt solely describes the overall image content without providing specific instructions for editing the designated region.
    \item The prompt is overly complex, containing unfeasible editing requirements or specialized terminology that renders it incomprehensible or unactionable.
    \item The prompt exhibits significant grammatical errors, rendering it unintelligible.
    \item The edited image is identical to the original image, indicating no effective modification.
    \item The image content violates ethical standards or induces severe discomfort and displeasure in viewers.
\end{itemize}

Additionally, if a portion of the prompt pertains to the subject within the outlined region while another portion refers to areas outside the region, only the part of the prompt corresponding to the outlined subject should be considered for scoring. All evaluations should be based solely on the relevant portion of the prompt, and such prompts should be subsequently corrected. This scenario differs from the previously mentioned cases and does not necessitate skipping the scoring process.

\subsection{Detailed Human Scoring Criteria}

The \textbf{overall editing quality} scoring range is $1$-$5$, categorized into five levels: bad, poor, fair, good, and excellent. This score should comprehensively consider both \textbf{adherence to the prompts} and the \textbf{presentation quality of the editing}, with higher priority given to instruction adherence. Cases where the instructions are completely or partially disregarded should receive low scores (\textbf{limited to 1-2 points}). The evaluation of instruction adherence should be lenient, meaning that as long as the editing requirements for the subject within the outlined region are fulfilled, the prompt is considered followed. The reference criteria for the five levels are as follows:

\begin{itemize}
    \item \textbf{Excellent (5)} The edited region fully adheres to the instruction requirements, and the overall image editing appears natural and harmonious.
    \item \textbf{Good (4)} The edited region fully adheres to the instruction requirements, and the overall image editing is reasonably natural and harmonious.
    \item \textbf{Fair (3)} The edited region adheres to the instruction requirements, but the overall image editing exhibits moderate naturalness and harmony.
    \item \textbf{Poor (2)} The edited region adheres to the instruction requirements, but the overall image editing is suboptimal; or the edited region does not fully adhere to the instruction requirements.
    \item \textbf{Bad (1)} The edited region adheres to the instruction requirements, but the overall image editing is highly unnatural and discordant; or the edited region does not fully adhere to the instruction requirements.
\end{itemize}

The \textbf{prompt completion} scoring range is 1-3, categorized into three levels: non-completion, partial completion, and full completion. The reference criteria for the three levels are as follows. Note that the evaluation of prompt completion generally does not consider the editing effect, focusing solely on whether the prompt is fulfilled (unless the prompt explicitly includes requirements for the editing effect, in which case the editing effect must also be considered).
\begin{itemize}
    \item \textbf{Full Completion (3)} The edited region accurately and completely follows the text description.  
    \item \textbf{Partial Completion (2)} The edited region partially follows the text description.  
    \item \textbf{Non-Completion (1)} The edited region does not follow the text description at all.  
\end{itemize}

The scoring range for \textbf{harmony} is 1-5, categorized into five levels: bad, poor, fair, good, and excellent. \textbf{Harmony evaluation is independent of the prompt} and is based solely on the visual assessment of the image. Specifically, it considers the consistency and harmony of the edited local region with the overall style and semantics of the image, focusing more on incongruity or semantic confusion rather than factual consistency. The reference criteria for the five levels are as follows:  

\begin{itemize}
    \item \textbf{Excellent (5)} The edited region is highly harmonious with the surrounding areas, with fully consistent spatial logic and details, and extremely smooth boundaries.  
    \item \textbf{Good (4)} The edited region is highly harmonious with the surrounding areas, with consistent spatial relationships and relatively smooth boundaries.  
    \item \textbf{Fair (3)} The edited region is moderately harmonious with the surrounding areas, with generally consistent spatial relationships but some discrepancies, and slight editing traces at the boundaries.  
    \item \textbf{Poor (2)} The edited region is poorly harmonious with the surrounding areas, with noticeable inconsistencies in spatial relationships and obvious editing traces at the boundaries.  
    \item \textbf{Bad (1)} The edited region is extremely discordant with the surrounding areas, with significant inconsistencies in spatial relationships and highly noticeable editing traces at the boundaries.  
\end{itemize}

The scoring range for \textbf{naturalness of the edited region} is 1-5, categorized into five levels: bad, poor, fair, good, and excellent. \textbf{The evaluation of naturalness is independent of the prompt} and is based solely on the edited area, specifically the outlined region. The reference criteria for the five levels are as follows:  

\begin{itemize}
    \item \textbf{Excellent (5)} The edited region appears highly natural, with excellent visual presentation.  
    \item \textbf{Good (4)} The edited region appears quite natural, with good visual presentation.  
    \item \textbf{Fair (3)} The edited region appears moderately natural, with acceptable visual presentation.  
    \item \textbf{Poor (2)} The edited region appears unnatural, with subpar visual presentation.  
    \item \textbf{Bad (1)} The edited region appears highly unnatural, with poor visual presentation, often appearing distorted or bizarre.  
\end{itemize}

% \section{Supplementary Details for Editing Tools}
% Our EPAIQA dataset encompasses a total of 12 distinct editing tools, whose corresponding usage references are provided as follows:
% \begin{itemize}
%     \item \textit{\textbf{Stable Diffusion}}: {https://platform.stability.ai/}
%     \item \textit{\textbf{Dalle-2}}: {https://openai.com/index/dall-e-2/}
%     \item \textit{\textbf{Midjourney}}: \href{https://www.midjourney.com/}{\textit{Midjourney}}
%     \item \href{https://github.com/cientgu/InstructDiffusion}{\textit{InstructDiffusion}}
%     \item \href{https://github.com/ml-research/ledits_pp}{\textit{Ledits++}}
%     \item \href{https://huggingface.co/diffusers/sdxl-instructpix2pix-768}{\textit{InstructPix2Pix}}
%     \item \href{https://github.com/cvlab-kaist/Perturbed-Attention-Guidance}{\textit{Perturbed-Attention Guidance (PAG)}}
%     \item \href{https://github.com/ruilin19/DiffEdit-by-Stable-Diffusion}{\textit{DiffEdit}}
%     \item \href{https://huggingface.co/stabilityai/stable-diffusion-2-inpainting}{\textit{Stable Diffusion v2 Inpainting}}
%     \item \href{https://huggingface.co/diffusers/stable-diffusion-xl-1.0-inpainting-0.1}{\textit{SD-XL Inpainting 0.1}}
%     \item \href{https://github.com/alimama-creative/FLUX-Controlnet-Inpainting}{\textit{FLUX-Controlnet-Inpainting}}
%     \item \href{https://huggingface.co/TTPlanet/HunyuanDiT_Controlnet_inpainting}{\textit{HunyuanDiT}}
% \end{itemize}

\section{Prompts for Dataset Construction}
\subsection{Prompts for Local Subject Extraction and Bounding}
\label{Gemini}
We utilize prompts provided to \textit{Gemini} to extract the primary subject within the image and ensure its uniqueness, thereby facilitating the subsequent outlining of the subject.

\noindent \textbf{\textit{Prompts:}} \textit{What is the main object in this picture? Please describe it in one word. How many main objects are there in each image? When outputting, please separate the word description and quantity with | and without space, and the maximum quantity is 10.}

For responses in the format \textit{$\langle subject | number \rangle$} generated by \textit{Gemini}, we first filter based on the number parameter, retaining only images where the subject count is exactly one. Subsequently, the corresponding subject and image are input into \textit{Dino-V1} to perform subject localization and bounding.

\subsection{Prompts for Editing Prompt Generation}
\label{editing prompt}
We classify local editing tasks into two distinct levels of difficulty, each associated with different models. For \textbf{complex tasks}, we execute four types of editing operations: object enhancement, object operation, semantic change, and style change. For \textbf{simple tasks}, we limit the editing operations to object operation and style change. Additionally, we leverage \textit{GPT-4o} to generate local editing prompts, which are subsequently utilized as inputs for our editing models.

\noindent \textbf{\textit{Prompts for complex task:}} \textit{Generate a prompt specially for image editing type: \textbf{[type]} for the subject in the frame. The prompt must meet the following  criteria:
\begin{enumerate}
    \item Target Focus: The edit command must only apply to the main subject within the specified bounding box.
    \item Single Action Rule: Each prompt should include only one editing command and with only one simple editing action related to {type} per prompt.
    \item Simplicity: Use simple, easy-to-understand language. Avoid complex artistic, aesthetic, or material-specific terminology.
    \item Semantic Consistency: Ensure that after the edit, the subject's semantic or identity remains largely similar to the original. For example, if replacing a dog, acceptable alternatives could be a cat or another small quadrupedal mammal, but not a human or an inanimate object.
    \item Overall Image Harmony: The editing should keep the overall image semantically coherent. For instance, when modifying a face, instructions may specify a change to a ceramic-smooth texture, but avoid material changes (like wood) that would create noticeable discordance in the image.
    \item Generated prompt cannot contain box information.
    \item The prompt must be a concise sentence with only the action instruction itself and without extra prefix or suffix.
\end{enumerate}}

\noindent \textbf{\textit{Prompts for simple tasks:}} \textit{Generate a prompt in the form of a noun specially for image editing type: \textbf{[type]} for the subject in the frame. The prompt must meet the following  criteria:
\begin{enumerate}
    \item Generate prompts is edited results in the form of 'a' + noun, noun no more than one words.
    \item Focus only on the main object within the bounding box.
    \item Each prompt must contain only one editing command for the main object.
    \item Ensure the edited object retains a similar semantic meaning to the original (e.g., replace a dog with a cat or another breed, but not a human or inanimate object).
    \item Ensure the edited object maintains semantic harmony with the overall image (e.g., 'a porcelain-like face' is acceptable, but avoid unrealistic materials like wood).
    \item Generated prompt cannot contain box information.
    \item Only generate one prompt at a time.
    \item The content in prompt must be different from the original object in the box.
\end{enumerate}}

\subsection{Prompts for Prompt Cleaning}
For certain prompts (e.g., those targeting multiple subjects while only one is within the outlined region), a cleaning process is required. We input both the original prompt and the cleaning prompt into \textit{GPT-4o} to obtain the refined and validated version.

\noindent \textbf{\textit{Prompts for cleaning:}} \textit{I have one picture with the boxed object and one corresponding image editing prompt: \textbf{[prompt]}, which is used to edit the object in the box, but sometimes the prompt does not meet the requirements. Now please analyze the picture and prompt and modify the prompt according to the following suggestions:
\begin{enumerate}
    \item Analyze each prompt and identify the boxed object in the image.
    \item If the prompt requests editing multiple objects with different types, discard the objects outside the bounding box and keep only the editing instruct corresponding to the main object within the bounding box.
    \item If the prompt requests editing multiple objects of the same type (like two birds of the same species) but only one of them is boxed, please modify the prompt to focus on the boxed object  by adding detailed semantic descriptions to highlight the boxed object (i.e. the bird in the left).
    \item For prompts which do not in the upper cases, please output the original prompt without any modification.
    \item Ensure the modified prompt applies the same editing action as the original given prompt but only to the boxed object.
    \item Output the modified prompts in a clear and concise format.
    \item Please simulate as if you are watching a picture without the bounding box, so do not mention any information related to the bounding box.
    \item The prompt must be a concise sentence which is the result prompt itself and without extra prefix or suffix.
\end{enumerate}}

\section{Prompts for Model Training}
\label{modeltra}
\subsection{Prompts for Grounding Pre-training}
\label{pretrain}

We train the grounding capability of \textit{GPT-4o} by providing it with \textbf{question-answer pairs} as prompt inputs. Below are the detailed question-answer pairs utilized for this purpose.
\begin{itemize}
    \item \textbf{\textit{Question:}} \textit{The original image i: \textbf{[image]}, the image after partial editing is: \textbf{[image]}. Please output the representation of the local editing area. It is expressed as four decimals in the range [0,1] (as a percentage of the image), where the first two digits are the horizontal/vertical coordinates of the center point of the editing area, and the last two digits are the width/height coordinates of the size of the editing area}
    \item \textbf{\textit{Answer:}} \textit{The four coefficients representing the editing area are: $\langle x_{\text{true}}, y_{\text{true}}, z_{1_{\text{true}}}, z_{2_{\text{true}}} \rangle$.}
\end{itemize}

\subsection{Prompts for Quantitative Prediction Training}
\label{stage2}
Initially, we generate Chain-of-Thought reasoning concerning local naturalness and harmony as foundational prior knowledge. The prompt employed for generating the CoT is as follows.

\noindent \textbf{\textit{Prompts for CoTs (Harmony) :}} \textit{Now there is a task of partial editing of an image. The edited image is: \textbf{[image]}. This image is considered having a lower degree of harmony between the edited area and the overall image, based on the given edited image, please analyze from the perspective of the edited image itself why the image editing having a lower degree of harmony between the edited area and the overall image. Please describe your analysis results in concise language, within two sentences.}

\noindent \textbf{\textit{Prompts for CoTs (Local Naturalness) :}} \textit{Now there is a task of partial editing of an image. The edited area is: \textbf{[image]}. This image is considered having a lower degree of naturalness in the edited area, based on the given edited area, please analyze from the perspective of the edited area itself why the image editing having a lower degree of naturalness in the edited area. Please describe your analysis results in concise language, within two sentences.}

Subsequently, we enhance capability of \textit{GPT-4o} by incorporating CoT reasoning as prior knowledge, utilizing question-answer pairs as instructional inputs to train its ability to predict local naturalness and harmony. The specific prompts are outlined below.

\begin{itemize}
    \item \textbf{\textit{Question (Harmony) :}} \textit{The image after partial editing is: \textbf{[image]}. \textbf{[CoT]}. Please rate the harmony between the edited area and the overall edited image (There are 5 levels in total: bad, poor, fair, good, excellent). Then output the harmony level.}
    \item \textit{\textbf{Answer (Harmony) :}} \textit{Harmony level is: \textbf{[harmony level]}.}
\end{itemize}

\begin{itemize}
    \item \textbf{\textit{Question (Naturalness) :}} \textit{The editing area of image after partial editing is: \textbf{[image]}. \textbf{[CoT]}. Please rate the naturalness of the edited area (There are 5 levels in total: bad, poor, fair, good, excellent). Then output the naturalness level.}
    \item \textit{\textbf{Answer (Naturalness) :}} \textit{Naturalness level is: \textbf{[naturalness level]}.}
\end{itemize}

\subsection{Prompts for Overall Editing Quality Assessment Training}
\label{COT}

Initially, we generate CoT reasoning to evaluate the overall editing quality. The analysis strictly follows a structured process: first assessing the degree of prompt adherence, then evaluating harmony and local naturalness, and finally deriving the overall editing effectiveness. Several specific examples of the generation prompt is provided below.

\begin{itemize}
    \item \textbf{\textit{Prompts for prompt completion:}} \textit{This image is considered following prompt at all, based on the given original image, the edited image and the editing instructions, please analyze from the perspective of the edited image itself why the image editing following prompt at all. Please describe your analysis results in concise language, within two sentences.}
    \item \textbf{\textit{Prompts for local naturalness:}} \textit{This image is considered having a lower degree of naturalness in the edit area, based on the given original image, the edited image and the editing instructions, please analyze from the perspective of the edited image itself why the image editing having a lower degree of naturalness in the edit area. Please describe your analysis results in concise language, within two sentences.}
\end{itemize}

\begin{figure}[htbp]
    \centering
    \begin{subfigure}[b]{0.97\linewidth}     
        \includegraphics[width=\linewidth]{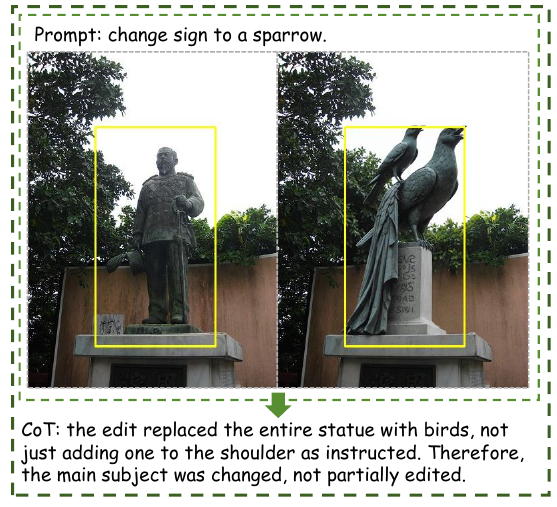}
        \caption{A CoT example for bad prompt completion.}
        % \label{fig:total2sub1}
    \end{subfigure}
    \begin{subfigure}[b]{0.96\linewidth}
        \centering
        \includegraphics[width=\linewidth]{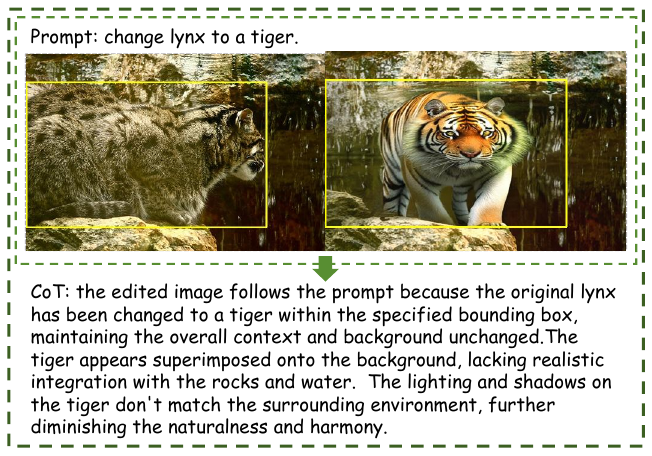}
        \caption{A CoT example for bad harmony and local naturalness.}
        % \label{fig:total2sub2}
    \end{subfigure}
    \begin{subfigure}[b]{0.95\linewidth}
        \centering
        \includegraphics[width=\linewidth]{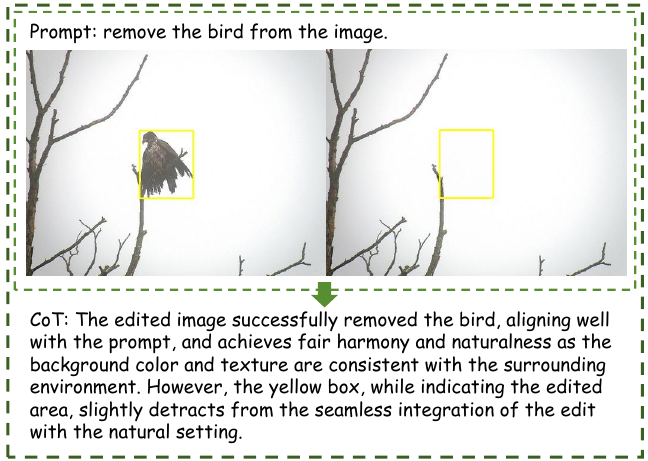}
        \caption{A CoT example for good editing quality.}
        % \label{fig:total2sub2}
    \end{subfigure}
    \vspace{-5pt}
    \caption{More information on data distribution}
    \label{fig:cot cases}
\end{figure}

We integrate the generated CoT reasoning (the examples are shown in Fig. \ref{fig:cot cases}) into the final training prompt for \textit{GPT-4o}, enabling it to predict and evaluate the overall editing quality while equipping it with the capability to provide explanatory feedback. The training prompts are structured in the form of question-answer pairs, as illustrated below.

\begin{itemize}
    \item \textit{\textbf{Question:}} \textit{The original image is: \textbf{[image]}, the image after partial editing is: \textbf{[image]}. The partial editing prompt is: \textbf{[prompt]}. First, determine whether the prompt is followed. If not, analyze it and directly give the overall quality level as bad. If partially followed, analyze it and directly give the overall quality level as poor. If followed, first analyze the prompt completion, the local naturalness of the edited area and the overall harmony, then please give the final overall editing quality level based on the analysis and the overall presentation quality of the edited image.}
    \item \textbf{\textit{Answer:}} \textit{The prompt completion is \textbf{[Completion Level]}. \textbf{[CoT of Prompt Completion]}. The harmony: \textbf{[CoT of Harmony]}. The local naturalness:\textbf{[CoT of Naturalness}]. Therefore, the overall editing quality level of the image is \textbf{[Overall Editing Level]}}
\end{itemize}

% \section{Key Patterns Visualization}

% We generate separate \textbf{word clouds} for the editing prompts and the CoTs to provide a visual representation of their respective key terms in Fig. \ref{fig:cloud}
% \begin{figure*}[htbp]
%     \centering
%     \begin{subfigure}[b]{0.48\linewidth}    
%         \centering
%         \includegraphics[width=0.95\linewidth]{samples/imgs/Prompt1.pdf}
%         \caption{Editing prompts}
%         \label{fig:cloud1}
%     \end{subfigure}
%     \begin{subfigure}[b]{0.48\linewidth}
%         \centering
%         \includegraphics[width=0.95\linewidth]{samples/imgs/CoT1.pdf}
%         \caption{CoTs}
%         \label{fig:cloud2}
%     \end{subfigure}
%     \vspace{-12pt}
%     \caption{Word clouds}
%     \label{fig:cloud}
% \end{figure*}
\section{Evaluation Experiments Supplementary Settings}
\label{Evaluation}
This section primarily supplements the experimental setup for the Stage-3 quality explanation in Sec. \ref{Experiments Setups}. First, the $300$ test data samples are categorized into three main types:

1. Type 1 (106 samples), where the edits either partially or completely fail to follow the prompts. The corresponding ground truth includes only the prompt completion analysis and overall quality level attribution. 

2. Type 2 (88 samples), where the edits follow the instructions, but at least one dimension of naturalness or harmony has a score below 3, and the overall score is also below 3. The corresponding ground truth includes prompt completion, naturalness, and harmony analyses, as well as overall quality level attribution. 

3. Type 3 (106 samples), where the edits follow the prompts, both naturalness and harmony are above 3, and the overall score is above 3. The corresponding ground truth also includes prompt completion, naturalness, and harmony analyses, as well as overall quality level attribution. 

The definitions of \textbf{PA}, \textbf{LNA}, \textbf{GHA}, and \textbf{overall sub-dimensions}, along with their GPT scoring criteria, are presented as follows:
\begin{itemize}
 \item  \textbf{\textit{PA}}: \textit{If the judgment in the  model response regarding whether the instructions are followed is consistent with the standard answer, 
                   and the analysis meaning is also consistent, assign 2 points. If the judgment is consistent, but there is a discrepancy in the analysis, assign 1 point. 
                   If the judgment differs, assign 0 points.}
        
\item \textbf{\textit{LNA}}: \textit{If the analysis regarding naturalness in the model response is basically consistent with the standard answer, 
                   or if both the answer and the standard answer do not include an analysis of naturalness, assign 2 points. 
                   If there is some difference, but the overall judgment is consistent, assign 1 point. 
                   If there is a complete inconsistency (opposite meaning), or the standard answer includes an analysis of naturalness that the answer does not, assign 0 points.}
        
\item \textbf{\textit{GHA}}:  \textit{If the analysis regarding harmony in the model response is basically consistent with the standard answer, 
                   or if neither the answer nor the standard answer includes an analysis of harmony, assign 2 points. 
                   If there is some difference, but the overall judgment is consistent, assign 1 point. 
                   If there is a complete inconsistency (opposite meaning), or the standard answer includes an analysis of harmony that the answer does not, assign 0 points.}
        
\item \textbf{\textit{Overall}}: \textit{If the overall editing quality level in the answer is essentially consistent with the standard answer 
                   (for example, `good' and `excellent,' or `bad' and `poor,' which are similar in meaning), and the overall summarization process is consistent, assign 2 points. 
                   If the overall editing effect level in the model response is essentially consistent with the standard answer, 
                   but there are differences in the process of summarizing the reasons for the editing effect level, assign 1 point. 
                   If the overall editing effect grade in the answer is inconsistent with the standard answer, assign 0 points.}
\end{itemize}                   
During testing, for each model's output on each test sample, we repeat the GPT scoring process five times, and the score for each dimension is the most frequent (in the case of a $2/2/1$ distribution, the highest score that appears twice is selected). Specifically, for Type 1 test samples, LNA and GHA are not included in the calculation. We ultimately use the average score of all test samples for each dimension as the final score for the model in that dimension.

\begin{figure*}[htbp]
  \centering
  \includegraphics[width=0.79\linewidth]{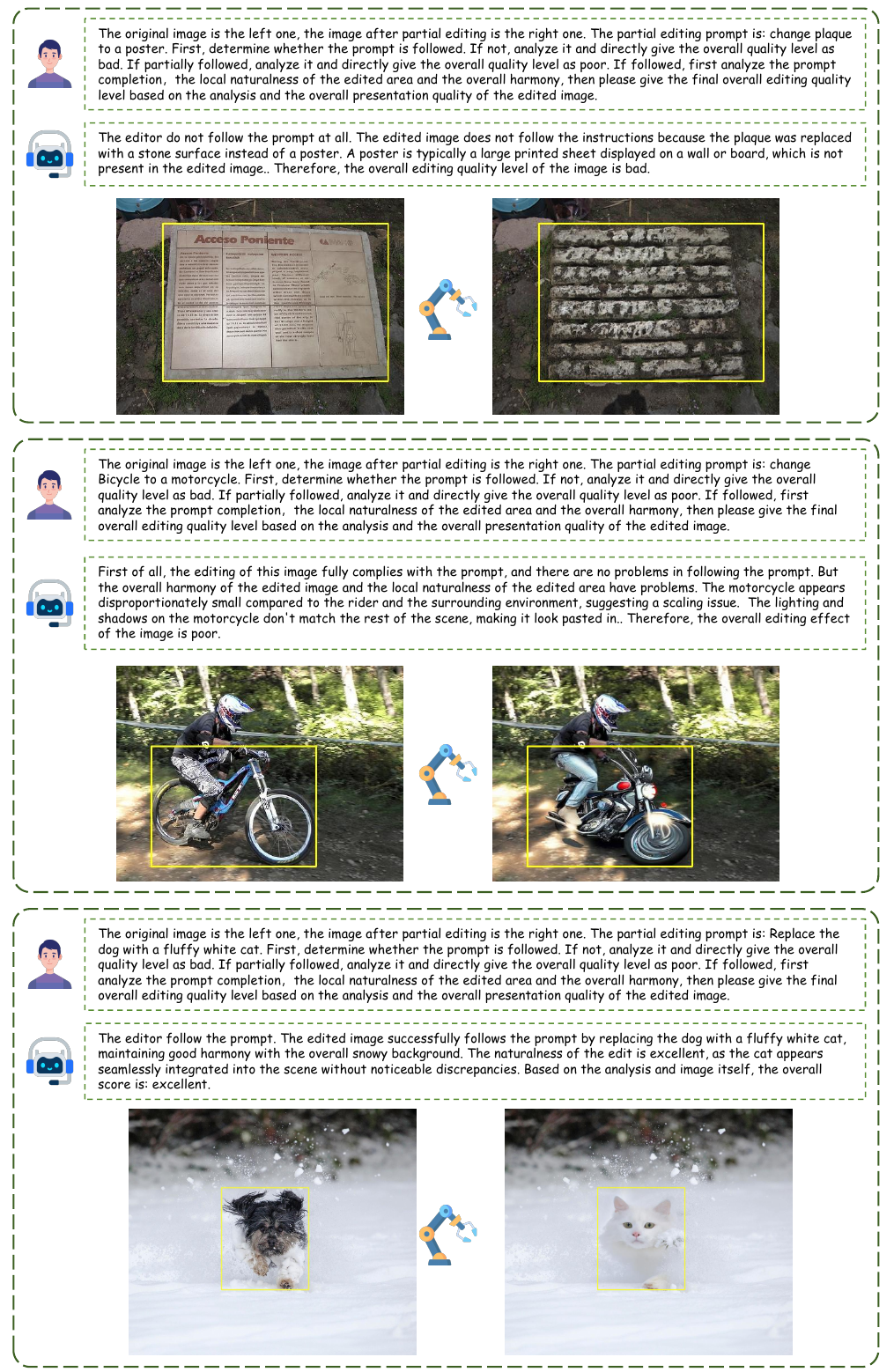}
  \vspace{-8pt}
  \caption{Case study examples of EPAIQA series models performance.}
   \label{case study}
\end{figure*}

\section{Case Studies}
\label{case}
We conducted a detailed case study on the \textbf{EPAIQA series models} to ensure that the models not only possess the capability to accurately assess the overall editing quality but also provide reliable analytical feedback with a coherent chain of thought. We selected three representative examples to demonstrate the model's performance (as shown in the Fig. \ref{case study}), which are categorized into the following three types:  
\begin{itemize}  
    \item \textbf{Failure to follow instructions, resulting in poor overall editing quality} (top section of Fig. \ref{case study});  
    \item \textbf{Following instructions but with suboptimal harmony or naturalness (or both), leading to poor overall editing quality} (middle section of Fig. \ref{case study});  
    \item \textbf{Following instructions with both harmony and naturalness well maintained, resulting in good overall editing performance} (bottom section of Fig. \ref{case study}).  
\end{itemize}

% \section{Limitations}
% We sincerely thank all the recruited volunteers for their valuable subjective feedback. Special thanks to our advisor for their guidance and to all collaborators for their support, which are essential to the completion of this work.
\section{Limitation}
Although our approach using LMM has shown significant advantages in scoring harmony, naturalness, and overall quality, its overall performance still falls short compared to the general testing performance of NSI-IQA and T2I-AGIQA. This suggests that the evaluation tasks for each dimension of PAIQA are inherently more challenging. There remains considerable room for further development, both in terms of dataset expansion and refinement, as well as the construction of specially designed models. Our current work lays a solid foundation for future expansion, and the tasks outlined above represent important potential directions for our next phase of exploration.
\section{License}
We are committed to releasing the core components of the \textbf{EPAIQA-15K dataset}, including all original images, edited images, and corresponding editing prompts, to facilitate research reproducibility. For the human-rated  data, we will release a carefully organized and anonymized portion publicly while retaining a private subset for future research and validation purposes. This approach ensures both transparency and the continued advancement of our work. We hope our endeavour will support academic progress in image editing and IQA fields while preserving valuable resources for ongoing exploration.

\end{document}